\documentclass{article}

\usepackage{arxiv}

\usepackage[utf8]{inputenc} 
\usepackage[T1]{fontenc}    
\usepackage{hyperref}       
\usepackage{url}            
\usepackage{booktabs}       
\usepackage{amsfonts}       
\usepackage{nicefrac}       
\usepackage{microtype}      
\usepackage{amsmath}        
\usepackage{lipsum}
\usepackage{graphicx}
\usepackage{xcolor}
\usepackage{multirow}
\usepackage{natbib}
\graphicspath{ {./images/} }
\usepackage{cleveref}
\usepackage{tabularx}
\usepackage{array}
\usepackage{pifont}
\usepackage{authblk}
\usepackage[table,xcdraw]{xcolor}
\usepackage[utf8]{inputenc}

\newcolumntype{L}[1]{>{\raggedright\arraybackslash}m{#1}}
\newcolumntype{C}[1]{>{\centering\arraybackslash}m{#1}}

\definecolor{precolor}{RGB}{248, 202, 198}   
\definecolor{postcolor}{RGB}{224, 224, 150}
\definecolor{taskcolor}{RGB}{194, 224, 244}  

\newcommand{\cmark}{\ding{51}}
\newcommand{\xmark}{\ding{55}}

\title{Do World Action Models Generalize Better than VLAs? A Robustness Study}
\author{
Zhanguang Zhang$^{1}$\thanks{Corresponding to \{zhanguang.zhang, yingxue.zhang\}@huawei.com}, Zhiyuan Li$^{1,2}$\thanks{work done during internship at Huawei Canada}, Behnam Rahmati$^{1}$, Rui Heng Yang$^{1}$, Yintao Ma$^{1}$, Amir Rasouli$^{1}$\\
\vspace{-0.14cm}
Sajjad Pakdamansavoji$^{1}$, Yangzheng Wu$^{1}$, Lingfeng Zhang$^{1}$, Tongtong Cao$^{1}$, Feng Wen$^{1}$\\
\vspace{0.1cm}
Xinyu Wang$^{1}$, Xingyue Quan$^{1}$ and Yingxue Zhang$^{1}$$^*$\\
\vspace{0.5cm}
$^{1}$Huawei Technologies \quad $^{2}$University of Toronto
}

\begin{document}

\maketitle

\begin{abstract}
Robot action planning in the real world is challenging as it requires not only understanding the current state of the environment but also predicting how it will evolve in response to actions. Vision-language-action (VLA), which repurpose large-scale vision-language models for robot action generation using action experts, have achieved notable success across a variety of robotic tasks. Nevertheless, their performance remains constrained by the scope of their training data, exhibiting limited generalization to unseen scenarios and vulnerability to diverse contextual perturbations. 
More recently, world models have been revisited as an alternative to VLAs. These models, referred to as world action models (WAMs), are built upon world models that are trained on large corpora of video data to predict future visual states. With minor adaptations, their latent representation can be decoded into robot actions. It has been suggested that their explicit dynamic prediction capacity, combined with spatiotemporal priors acquired from web-scale video pre-training, enables WAMs to generalize more effectively than VLAs. In this paper, we conduct a comparative study of prominent state-of-the-art VLA policies and recently released WAMs. We evaluate their performance on the LIBERO-Plus and RoboTwin 2.0-Plus benchmarks under various visual and language perturbations. Our results show that WAMs achieve strong robustness, with LingBot-VA reaching 74.2\% success rate on RoboTwin 2.0-Plus and Cosmos-Policy achieving 82.2\% on LIBERO-Plus. While VLAs such as $\pi_{0.5}$ can achieve comparable robustness on certain tasks, they typically require extensive training with diverse robotic datasets and varied learning objectives. Hybrid approaches that partially incorporate video-based dynamic learning exhibit intermediate robustness, highlighting the importance of how video priors are integrated. Overall, our findings shed light on critical insights into the strengths of WAMs relative to VLAs, as well as the remaining challenges for real-world deployment. The evaluation code for the RoboTwin2.0-Plus benchmark is available at: https://robot-robustness.github.io/RoboTwin2.0-Plus/
\end{abstract}

\section{Introduction}
Robot motion planning in real-world environments—whether for navigation, manipulation, or locomotion—remains highly challenging. A key difficulty arises from the diversity and uncertainty of real-world settings, which makes it hard for robot policies to anticipate the consequences of their actions. Recently, vision–language–action (VLA) policies~\citep{pi05,openvla2025} have emerged as a new paradigm compared to traditional motion planning approaches~\citep{elbanhawi2014sampling}. By leveraging foundation models trained on large-scale vision and language data, VLA policies have demonstrated strong performance across a wide range of robotic tasks, including navigation~\citep{xu2024mobility}, manipulation~\citep{zheng2026xvla}, and locomotion~\citep{jiang2026wholebodyvla}.

Despite their strong performance, VLAs exhibit several notable limitations, including limited generalization ability~\citep{ma2026generalvla} and a lack of robustness to distractions and cluttered environments~\citep{rasouli2025distracted}. In essence, they often lack a fundamental understanding of the physical world that is necessary for consistent planning across diverse environments. To address this issue, recent work has begun incorporating world models into robotic policies. Traditionally, world models have been used primarily as simulators for model training and evaluation~\citep{cutler2015real,ha2018_worldmodels,hafner2020_dreamer}. More recently, they have been increasingly integrated into robot policies—including VLAs—in a variety of ways, such as auxiliary training objectives~\citep{Chen_2025_ICCV,cen2025worldvla}, planning modules~\citep{yin2025womap,gao2025adaworld}, or guidance mechanisms for flow-based policies~\citep{du2025dynaguide,huang2025ladiwm,chen2026h}. Building on this trend, new approaches go a step further by proposing the use of world models directly as control policies~\citep{kim2026cosmos,goswami2025osviwm, liao2025genie, li2026causal, ye2026dreamzero}, where the model’s latent representations are decoded into actions.
 
Given the many similarities between foundation models and world models, there is ongoing debate about the primary advantages of world models and whether their explicit use in planning is necessary, since foundation-based robot policies may already implicitly model the world dynamics. To investigate this question, we conduct a comparative study of state-of-the-art VLA policies and world action models (WAMs), aiming to highlight their differences under a range of contextual perturbations. More specifically, we leverage two enhanced manipulation benchmarks to evaluate the robustness of policy models: \textbf{LIBERO-Plus}~\citep{fei25libero-plus}, which introduces seven types of perturbations to single-arm manipulation tasks, and \textbf{RoboTwin 2.0-Plus}, an in-house benchmark that follows a similar perturbation protocol in the two-arm Aloha-Agilex setup of RoboTwin 2.0~\citep{chen2025robotwin}. 

Our findings reveal that WAMs generally demonstrate strong robustness to noise, lighting and layout perturbations in both single-arm and bimanual settings. This robustness is believed to be at least partially attributable to the spatiotemporal priors inherited from their world model backbones. While classic VLAs like $\pi_{0.5}$~\citep{pi05}, as well as hybrid approaches like MOTUS~\citep{bi2025motus} and VLA-JEPA~\citep{sun2026vla}, can achieve comparable or even superior robustness, they often require carefully curated and diverse datasets and/or explicit dynamic prediction objectives during the embodied pre-training phase. In contrast, the simplicity of the embodied pre-training phase represents a key advantage of WAMs over classic VLAs. However, the high inference overhead of WAMs remains a major challenge that limits their deployment in real-world robotic systems, with a single inference step being at least 4.8 times slower than $\pi_{0.5}$. Recent GigaWorld-Policy~\citep{ye2026gigaworld} and Fast-WAM~\citep{yuan2026fast} mitigate this issue by generating actions only at inference time, while jointly predicting future visual state or conditioning state prediction on actions during training. In our evaluation, Fast-WAM achieves performance on RoboTwin~2.0-Plus that is essentially on par with the state-of-the-art IDM-based WAM (LingBot-VA), but its robustness on LIBERO-Plus collapses sharply when the training data lacks diversity, suggesting that joint-denoising designs may rely more heavily on training-data diversity for generalization than IDM-style designs that explicitly condition action on a predicted future state. Their inference latency also remains substantial compared to the $\pi$-series, posing challenges for smooth real-world deployment. Further research is needed to more efficiently exploit the dynamic priors of world-model backbones and improve both training and inference efficiency.

\section{Related Works}

\subsection{Vision Language Action (VLAs)}
Recent progress in robot foundation models has focused on integrating perception, language understanding, and control within unified multimodal architectures. A central direction in this line of work is the development of vision–language–action (VLA) models that connect high-level semantic reasoning with low-level robot execution. Early work such as PaLM-E~\citep{driess2023palm} demonstrates that large language models can be adapted to embodied settings by representing visual observations and robot states as additional tokens in a transformer. This formulation shows that knowledge learned from internet-scale data can support downstream robotic manipulation and long-horizon tasks when paired with action prediction. Building on this paradigm, a series of works explored large-scale end-to-end VLA policies trained on robot demonstrations. RT-1 and RT-2~\citep{brohan2023rt1, zitkovich2023rt2} introduced transformer policies that map visual observations and language instructions directly to robot actions, showing that large multi-task datasets enable generalization across manipulation behaviors. RT-H~\citep{belkhale2024rth} further explored hierarchical policy structures that improve data efficiency and long-horizon execution. In parallel, the $\pi_0$ series~\citep{black2025pi0} studied scaling behavior in VLA policies, emphasizing the role of action representation and dataset diversity in enabling robust multi-task generalization.

A second line of work emphasizes data-centric scaling and cross-embodiment generalization. Octo~\citep{ghosh2024octo}, trained on the Open X-Embodiment dataset~\citep{o2024open}, demonstrated that a single policy can learn from heterogeneous datasets spanning multiple robots and tasks. OpenVLA~\citep{openvla2025} further advanced this direction by providing an open-source VLA implementation built on pretrained multimodal encoders, highlighting the importance of reproducible pipelines and shared datasets for scaling robot foundation models. More recent research has focused on improving reasoning and adaptation within VLA architectures. CoT-VLA~\citep{zhao2025cotvla} introduces visual chain-of-thought reasoning by predicting intermediate visual goals before generating actions, improving long-horizon task performance. ChatVLA-2~\citep{zhou2025chatvla2} explores integrating open-world reasoning capabilities from pretrained vision–language models into robotic policies. Meanwhile, X-VLA~\citep{zheng2026xvla} investigates scalable cross-embodiment learning through embodiment-specific soft prompts, and SimpleVLA-RL~\citep{li2026simplevlarl} demonstrates that reinforcement learning applied after imitation pre-training can significantly improve robustness and long-horizon execution.

\subsection{World Model in Robotics}
World models learn the internal representation of the world and can predict future visual states conditioned on actions. Although the concept has long been studied~\citep{craik1967nature,sandage1988observational,ha2018world}, it has recently experienced a surge in interest and has been applied across a wide range of applications, including image retrieval~\citep{Yuanmin_2025_CVPR}, autonomous driving~\citep{Hassan_2025_CVPR,Zhao_2025_CVPR}, medical imaging~\citep{Yue_2025_CVPR,yang2025medical}, face generation~\citep{zheng2025learning}, and robotics locomotion~\citep{hao2025neural,wang2025disentangled}, navigation~\citep{bar2025navigation,yao2025navmorph}, object manipulation~\citep{zhen2025learning}.

In the context of robotics, world models have been used in several ways, including as simulators for training and evaluation~\citep{shang2025roboscape}, as auxiliary modules to boost planning policies~\citep{yin2025womap}, or as policies themselves after certain adaptations~\citep{goswami2025osvi}. Below we provide a brief review of each of these categories. 

\subsubsection{World Models as Learned Simulators}

In robotics and embodied AI, world models are most operationally useful when treated as \textbf{learned simulators}: action-conditioned generative models that produce counterfactual futures for decision-making. Rather than focusing on perceptual reconstruction or open-ended generation, this view treats the model as a computational substrate for control, where imagined rollouts must remain action-grounded, decision-relevant, and stable under closed-loop distribution shift~\citep{ha2018world,hafner2019_planet,hafner2020_dreamer}. Importantly, simulation does not need to occur in pixel space. Latent dynamics models—such as recurrent state-space models and predictive state representations—enable compact rollouts as long as task-relevant structure is preserved~\citep{littman2001_psr}. This shift from reconstruction fidelity toward control-aligned prediction exposes a key tension: improvements in likelihood or visual fidelity do not necessarily translate to better planning or policy learning, especially under long-horizon drift or objective mismatch~\citep{lambert2020_objective_mismatch,soh2026_action_hallucination}.

Learned simulators are used in two main ways. First, they support explicit planning, where policies optimize actions by querying the model at test time. Latent planners such as PlaNet, PETS, and MBPO perform model predictive control with varying strategies to mitigate model error~\citep{hafner2019_planet,chua2018_pets,janner2019_mbpo}. More recently, V-JEPA~2-AC enables planning from image goals in latent space after large-scale pre-training on video data~\citep{assran2025vjepa2}. H-WM further explores hierarchical planning by coupling symbolic task-level prediction with visual state prediction, using intermediate subgoals to guide long-horizon robotic task and motion planning~\citep{chen2026h}.
Video-based and foundation-scale simulators extend this idea to high-dimensional observations, enabling goal-directed planning from images or large-scale video pre-training~\citep{finn2017_visualforesight,ebert2018_visualforesight}.  NVIDIA presents Cosmos-Predict2.5 as a multimodal world-generation model designed to support robotics planning and large-scale synthetic rollouts~\citep{nvidia2025_cosmos,ali2025_cosmospredict25}. 
Second, world models enable imagination-based policy optimization, where policies are trained directly on simulated trajectories, as in World Models, Dreamer, and its successors~\citep{ha2018_worldmodels,hafner2020_dreamer,hafner2023_dreamerv3}. Hybrid approaches such as TD-MPC combine short-horizon model rollouts with value bootstrapping to limit compounding error while retaining sample efficiency~\citep{hansen2022_tdmpc,hansen2024_tdmpc2}. Across these paradigms, the simulator’s value lies less in perceptual realism and more in producing predictions that are reliable for control.

In visual navigation, world models are increasingly used to support tasks such as image-goal navigation. The Navigation World Model (NWM)~\citep{bar2025navigation} enables goal-conditioned navigation by using a Model Predictive Control (MPC)~\citep{kouvaritakis2016model} framework to simulate and evaluate candidate trajectories. Building on this idea, a lightweight one-step world model~\citep{shen2026efficient} employs a 3D U-Net with spatial–temporal attention to predict future observations, improving robustness and efficiency in multi-modal goal navigation tasks. More recently, MindJourney~\citep{yang2025mindjourney} leverages world models for test-time scaling to enhance a vision–language model’s understanding of 3D dynamics.

\begin{table}[t]
\caption{\textbf{Summary of world action models (WAMs)}. \textbf{MOT}: mixture-of-transformers, indicating that separate transformer backbones are used for video and action streams, with cross-modal interactions facilitated via attention at each layer. Note that the released LingBot-VA model adopts a unified transformer for both modalities, instead of the mixture-of-transformer architecture described in the paper. \textbf{Pretrain Free}: the method does not require task-agnostic embodied pre-training. \textbf{Causal Pred.}: action prediction is conditioned on the generated visual state, or vice versa. \textbf{AR Gen.}: auto-regressive generation.}
\label{tab:wam_compare}
\centering
\setlength{\tabcolsep}{3pt}

\resizebox{0.9\linewidth}{!}{
\begin{tabular}{L{3cm} C{1.5cm} C{3cm} C{1.5cm} C{1.5cm} C{1.5cm} C{1.5cm}}
\toprule
\textbf{Model} & \textbf{\#Params} & \textbf{Backbone} & \textbf{MOT} & \textbf{Pretrain Free} & \textbf{Causal Pred.} & \textbf{AR Gen.} \\
\midrule

VPP\newline
\citep{hu2025video}
& 1.5B
& Stable Video Diffusion (SVD)
& \xmark
& \xmark
& \cmark
& \xmark \\

GE-Act\newline
\citep{liao2025genie} 
& 2.2B 
& LTX-Video-2B
& \cmark 
& \xmark
& \xmark
& \cmark \\

Cosmos-Policy\newline
\citep{kim2026cosmos}
& 2B 
& Cosmos-Predict2-2B 
& \xmark 
& \cmark
& \xmark
& \xmark \\

LingBot-VA\newline
\citep{li2026causal}
& 5.3B 
& Wan2.2-5B 
& \cmark ?
& \xmark
& \cmark
& \cmark \\

DreamZero\newline
\citep{ye2026dreamzero}
& 14B 
& Wan2.1-14B 
& \xmark
& \xmark
& \xmark
& \cmark \\

GigaWorld-Policy\newline
\citep{ye2026gigaworld}
& >5B
& Wan2.2-5B
& \xmark
& \xmark
& \cmark
& \xmark \\

Fast-WAM\newline
\citep{yuan2026fast}
& 6B
& Wan2.2-5B
& \cmark
& \cmark
& \xmark
& \xmark \\

\bottomrule
\end{tabular}
}
\end{table}

\subsubsection{World Models as Auxiliary Tasks}
In this scheme, world knowledge is acquired by augmenting the policy with predictive capacities. This is achieved by introducing auxiliary prediction tasks~\citep{Chen_2025_ICCV,cen2025worldvla,zhang2025dreamvla,sun2026vla,wang2026unified} or by co-training the policy with a predictive model~\citep{wang2026unified,li2025unified}. In \citet{Chen_2025_ICCV}, a GPT-like architecture is proposed in which video content is converted into latent motion tokens that are incorporated with the text and the initial image observation. The model is first pre-trained to predict future video tokens and then finetuned with an action head to generate motions. WorldVLA~\citep{cen2025worldvla} enhances the VLA policy by adding image prediction component to generate future scenes given the observation and current action. DreamVLA~\citep{zhang2025dreamvla} augments the policy with three auxiliary predictive tasks: semantic, depth and dynamic prediction. In \citet{sun2026vla}, the authors employ a teacher–student training framework in which a predictive video encoder generates dynamic targets from future visual states, while the VLM backbone produces latent action representations using only the current observation. These two latent representations are then mapped and combined to form future states, which are learned through an alignment loss. The authors of \citet{wang2026unified} achieve multitasking by using interleaved representations that combine discrete vision, language, and action tokens as input. Unified World Model~\citep{zhu2025unified} integrates action and video diffusion processes within a unified transformer architecture, with independent diffusion timesteps governing each modality. Following this scheme, the model can learn a forward dynamics, an inverse dynamics, and a video generator. MOTUS~\citep{bi2025motus} introduces a Mixture-of-Transformers (MoT) architecture that unifies video and action generation, enabling training across five tasks with diverse data types. The model in \citet{li2025unified} learns a policy by decoupling video-action decoding. Via a masking operation in video and action space, the model is induced to forward and inverse dynamics, as well as planning capability. For the navigation task. UniWM~\cite{dong2025unified} integrates prediction and control within a unified multimodal framework, first predicting actions and then autoregressively reconstructing future visual observations, with the reconstruction loss serving as an auxiliary supervision for action prediction.

\subsubsection{World Action Models (WAMs)}
Despite the success of the VLAs aforementioned, they are mostly finetuned from VLM backbone models pretrained with an auto-regressive next-token prediction objective. While this language-centric pre-training enables models to capture high-level visual semantics, it often neglects the perception and prediction of fine-grained world dynamics that are essential for precise robotic control. With recent advances in action-conditioned video generation, a growing body of research has begun to explore adapting video generation models for policy learning~\citep{bi2025motus, hu2025video, kim2026cosmos, li2026causal, ye2026dreamzero, ye2026gigaworld, yuan2026fast}.  Video Prediction Policy (VPP) by \citet{hu2025video} is among the earliest efforts to repurpose a video generation backbone for robot action generation. Specifically, a video diffusion model is first pretrained on text-guided video prediction task, and then further adapted to generate robot actions with a diffusion policy head conditioned on visual features encoded by the video model. Empirical results show that the video pre-training stage is crucial for the observed performance improvements. Building on this paradigm, mimic-video~\citep{pai2025mimicvideo} leverages a language-conditioned video generation model (i.e., Cosmos-Predict2-2B~\citep{agarwal2025cosmos}, while retaining the same two-stage training scheme. It introduces a flow matching-based action decoder that is trained from scratch and serves as an inverse dynamic model (IDM). 
GE-Act~\citep{liao2025genie}, built on the Genie Envisioner (GE) platform, follows a similar strategy. Leveraging a pretrained video diffusion model backbone (i.e., LTX-Video-2B), it introduces a lightweight, flow-matching action decoder which maps video model encoded latent features to robot action trajectories. 
Despite the success of mimic-video and GE-Act, training an additional action decoder from scratch can disrupt the latent space learned by the video model and incurs extra training cost. Instead, Cosmos-policy~\citep{kim2026cosmos} minimally adapts the diffusion process of the video generation model Cosmos-Predict2, encoding the robot state, future image and value estimates directly as latent frames. With these lightweight architectural modifications, the model is finetuned under joint training objectives for policy, world model and value prediction. The resulting model supports both direct policy generation and model-based planning through its predicted future state and value estimates.  
LingBot-VA~\citep{li2026causal} and DreamZero~\citep{ye2026dreamzero} enhance causal reasoning capacity of video-based policy model by unifying future visual state prediction and action inference within an interleaved sequence and autoregressively generating future predictions conditioned on previous step's outputs. This formulation enables efficient KV-cache memory integration while enforcing causal consistency---both of which are crucial for long-horizon robotic tasks. While using different video generation backbones, both methods address the challenge of slow inference of video models for robot control. To improve real-time performance, they accelerate inference through techniques such as asynchronous inference pipeline and partial video denoising, among other optimizations. 

GigaWorld-Policy~\citep{ye2026gigaworld} and Fast-WAM~\citep{yuan2026fast} aim to mitigate the slow inference time of WAMs by treating video generation as optional at test time. GigaWorld-Policy conditions future visual state prediction on actions, whereas Fast-WAM jointly generate both modalities. Both designs enable test-time action prediction without explicitly generating visual state, as required in other IDM-based approaches (e.g., LingBot-VA), thereby significantly reducing the inference latency. However, the reported inference latencies of GigaWorld-Policy (360 ms) and Fast-WAM (190 ms) remain substantially higher than that of $\pi_0$~\citep{black2025pi0} on a consumer-level device (73 ms), leaving considerable room for improvement. Beyond inference speed, Fast-WAM also demonstrates great data efficiency in both simulation and real-world tasks by requiring only task-specific finetuning, without costly embodied pre-training on large-scale task-agnostic data. As shown in our evaluation (\Cref{tab:robotwin_plus}), Fast-WAM is essentially on par with the state-of-the-art LingBot-VA on RoboTwin 2.0, despite the absence of embodied pre-training---further substantiating the data efficiency of the WAM paradigm.

The key characteristics of recent WAMs are summarized in \Cref{tab:wam_compare}. We only consider the methods that leverage a pretrained world model backbone for robot action generation, with minimal or no architecture modifications. Consequently, MOTUS is excluded from this list: although it adopts a pretrained Wan2.2-5B for video generation, it relies on an additional VLM for action generation rather than the world model backbone itself. Typically, lightweight modifications are introduced to the video backbones to encode robot joint state and produce robot actions. While some approaches (e.g., Cosmos-Policy, GigaWorld-Policy) employ a unified transformer backbone for both video and action streams, others such as GE-Act and Fast-WAM adopt mixture-of-transformers (MOT) architectures, using a smaller, dedicated transformer for action modeling. Regarding training objectives, WAM models are typically trained to predict future visual state and actions, and in most cases require embodied pre-training on large-scale robot data. The design of causal attention also varies across methods. For instance, LingBot-VA conditions action generation on predicted visual state, whereas Cosmos-Policy and DreamZero jointly denoise both modalities. In contrast, GigaWorld-Policy conditions video generation on actions. Autoregressive generation is employed in GE-Act, LingBot-VA and DreamZero to condition prediction on historical context, improving temporal consistency and inference efficiency.

\begin{table*}[t]
\caption{\textbf{Training data of VLAs and WAMs across different training stages.} \textbf{Training stages}: \colorbox{precolor}{Embodied Pre-training}, \colorbox{postcolor}{Embodied Post-training}, \colorbox{taskcolor}{Task-specific Finetuning}. \textbf{PT}: filtered data for embodied post-training.}
\label{tab:training_data}
\centering
\small
\setlength{\tabcolsep}{4pt}
\begin{tabularx}{\linewidth}{lp{3.5cm}p{2.6cm}p{3.8cm}p{3cm}}
\toprule
\textbf{Model}
  & \textbf{Semantic Understanding \& Spatial Grounding}
  & \textbf{General Video}
  & \textbf{Robot Data}
  & \textbf{Task-specific Data} \\
\midrule

\multicolumn{5}{c}{\textit{VLAs}} \\

\arrayrulecolor{lightgray}
\midrule

$\pi_0$
  & -
  & -
  & \colorbox{precolor}{Cross-embodiment (>10k h)}
  & \colorbox{taskcolor}{Trajectories (5--100 h)} \\[4pt]

OpenVLA-OFT
  & -
  & -
  & \colorbox{precolor}{Cross-embodiment (970k)}
  & \colorbox{taskcolor}{Trajectories (20--300)} \\[4pt]

X-VLA
  & -
  & -
  & \colorbox{precolor}{Cross-embodiment (288k)}
  & \colorbox{taskcolor}{Trajectories (50)} \\[4pt]

$\pi_{0.5}$
  & \colorbox{precolor}{High-level planning} & -
  & \colorbox{precolor}{Mobile manipulation (400 h)}
  & \colorbox{taskcolor}{Trajectories (1--20 h)}
  \\
  & \colorbox{precolor}{Web data-VQA} & 
  & \colorbox{precolor}{Multi-env. tabletop} & 
  \\
  & \colorbox{precolor}{Web data-captioning} &
  & \colorbox{precolor}{Cross-embodiment} & 
  \\
  & \colorbox{precolor}{Web data-grounding} & & \colorbox{postcolor}{Mobile manipulation (PT)} & 
  \\
  & \colorbox{postcolor}{High-level planning (PT)} & & \colorbox{postcolor}{Multi-env.\ tabletop (PT)} & 
  \\
  & \colorbox{postcolor}{Verbal instruction} & & &
  \\
  & \colorbox{postcolor}{Web data (PT)} & & &  
  \\[4pt] 

\arrayrulecolor{black}
\midrule


\multicolumn{5}{c}{\textit{VLA + WM}} \\
\arrayrulecolor{lightgray}
\midrule

VLA-JEPA
  & -
  & \colorbox{precolor}{Human Ego (220k)}
  & \colorbox{precolor}{Single-embodiment (76k)}
  & \colorbox{taskcolor}{Trajectories (100)} \\[4pt]

MOTUS
  & -
  & \colorbox{precolor}{Human Ego (231k)}
  & \colorbox{precolor}{Cross-embodiment (781k)} 
  & \colorbox{taskcolor}{Trajectories (100)} \\
  & & & \colorbox{postcolor}{Cross-embodiment (781k)} \\
  & & & \colorbox{postcolor}{Task-agnostic data (1k)} \\ [4pt] 
  

\arrayrulecolor{black}
\midrule
\multicolumn{5}{c}{\textit{WAMs}} \\
\arrayrulecolor{lightgray}
\midrule

Cosmos-Policy
  & -
  & -
  & -
  & \colorbox{taskcolor}{Trajectories (185)} \\[4pt]

DreamZero
  & -
  & -
  & \colorbox{precolor}{Single-embodiment (500 h)}
  & \colorbox{taskcolor}{Trajectories (12--40 h)} \\

GE-Act
  & -
  & -
  & \colorbox{precolor}{Single-embodiment (3k h)}
  & \colorbox{taskcolor}{Trajectories (1 h)} \\[4pt]

LingBot-VA
  & -
  & -
  & \colorbox{precolor}{Cross-embodiment (16k h)}
  & \colorbox{taskcolor}{Trajectories (50)} \\[4pt]

GigaWorld-Policy
  & -
  & \colorbox{precolor}{Human Ego (4.5k h)}
  & \colorbox{precolor}{Cross-embodiment (6.7k h)}
  & \colorbox{taskcolor}{Trajectories (50)} \\[4pt]

Fast-WAM
  & -
  & -
  & -
  & \colorbox{taskcolor}{Trajectories (60 h)} \\[4pt]

\arrayrulecolor{black}
\bottomrule
\end{tabularx}
\end{table*}

\subsection{Differences between WAMs and VLAs} \label{sec:diff_wam}

WAMs differ from VLAs in their backbone model choices, training strategies, and prediction schemes. While WAMs leverage video generation models pretrained for video synthesis, VLAs are typically built upon VLM backbones pretrained for next-token prediction. The training datasets used by several influential VLAs, WAMs, and hybrid approaches are summarized in \Cref{tab:training_data}. In terms of notation, the training scheme is divided into two phases: \textbf{task-agnostic embodied pre-training} and \textbf{task-specific finetuning}. The embodied pre-training may be followed by a separate post-training stage, as in the case of $\pi_{0.5}$~\citep{pi05}. During the embodied pre-training stage, WAMs are commonly trained to jointly predict future visual states and actions using robot manipulation data. In contrast, VLAs are typically trained either for next-token prediction or action diffusion, often using not only robot data but also multi-modal web data and human videos in some cases~\citep{pi05, BuQ-RSS-25}. In terms of prediction schemes, VLAs typically map current state $h_t$ directly to the action $a_t$ via $p_\theta(a_t | h_t)$. In contrast, WAMs either jointly predict the future visual state $h_{t+1}$ and action $a_t$ via $p_\phi(h_{t+1}, a_t | h_t)$ or first predict the future visual state and then condition action generation on the predicted state, analogous to an inverse dynamic model (IDM), as $p_\phi(h_{t+1} | h_t) \cdot g_\psi(a_t | h_t, h_{t+1})$. In a less common causal formulation, GigaWorld-Policy predicts the action first, then generates future visual state conditioned on the action.

The pre-training stage of video generation backbones such as Cosmos-Predict2 involves future visual state prediction—specifically the $p_\phi(h_{t+1} \mid h_t)$ component—trained on diverse internet-scale video data covering natural dynamics, hand motion, driving, etc. This video pre-training objective enhances the model's ability to capture general physical dynamics and to predict fine-grained spatiotemporal state transitions. Given this prior physical knowledge, the embodied pre-training phase of WAMs can primarily focus on establishing general action prediction, $g_\psi(a_t \mid h_t, h_{t+1})$, which is a relatively easier learning problem. Prior work suggests that agents capable of generalizing to multi-step goal-directed tasks must effectively learn predictive structure in the environment~\citep{richens2025general}. In contrast, the backbones of VLAs---VLMs---are typically trained on static image-text data and therefore often lack fine-grained dynamic prediction capability. As a result, VLAs generally require more diverse geometric grounding, video data, and robotic data during embodied pre-training to implicitly acquire models of world dynamics.

\section{Experiments}

\subsection{Datasets}

To systematically compare the robustness of world action models (WAMs) and vision-language-action (VLA) models under diverse perturbation factors, we propose the \textbf{RoboTwin 2.0-Plus} benchmark, constructed upon RoboTwin 2.0~\citep{chen2025robotwin} and following the perturbation protocol of LIBERO-Plus~\citep{fei25libero-plus} with minor parameter adjustments. Specifically, perturbations are applied along the following axes: (1)~\textit{Camera}, altering the viewpoint pose of the third-person camera; (2)~\textit{Robot}, varying the initial joint configuration; (3)~\textit{Language}, paraphrasing or altering task instructions; (4)~\textit{Light}, modifying illumination intensity, shadow direction, and color temperature; (5)~\textit{Background}, changing table and scene textures; (6)~\textit{Noise}, applying photometric distortions to input images; and (7)~\textit{Layout}, introducing task-irrelevant distractive objects into the workspace. This benchmark is designed to facilitate evaluation of WAMs and VLAs that provide open-source checkpoints trained on RoboTwin 2.0. Further details are provided in~\Cref{app:robotwin_plus}. In addition, we also evaluate existing approaches on the open-sourced \textbf{LIBERO-Plus} benchmark~\cite{fei25libero-plus}. 

\begin{figure}[t]
    \centering
    \includegraphics[width=0.8
    \linewidth]{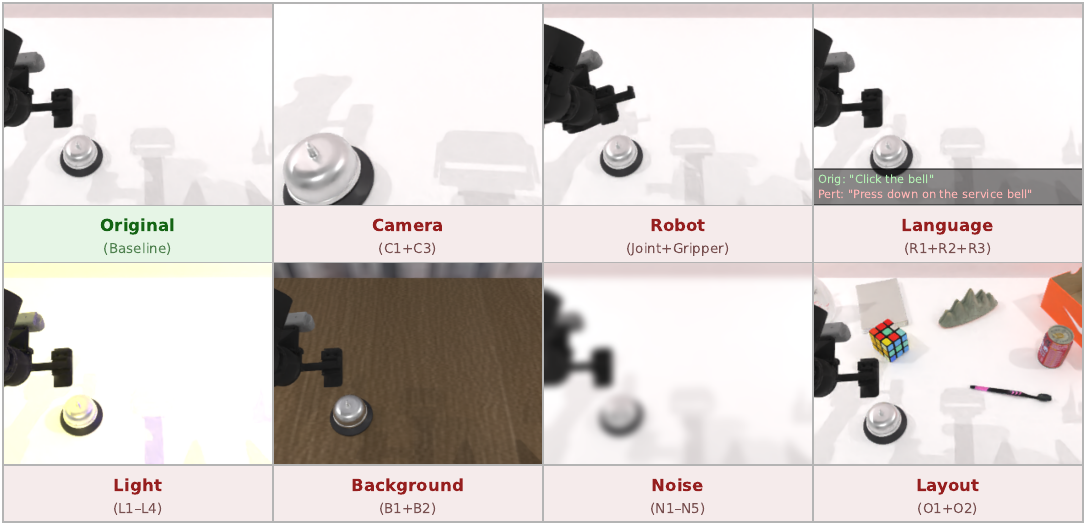}
    \caption{\textbf{Examples of perturbations on RoboTwin 2.0-Plus tasks.} Definition of notations (N1, L1, etc.) can be found in \Cref{app:robotwin_plus}}
    \label{fig:robotwin_plus}
\end{figure}

LIBERO-Plus and RoboTwin 2.0-Plus differ substantially in their underlying environments, as summarized in \Cref{tab:benchmark_comparison}. The primary distinction lies in their observation and action spaces. LIBERO-Plus employs a 7-DoF Franka Panda robot equipped with two cameras—one third-person view and one wrist-mounted—both operating at a resolution of $256\times256$. In contrast, RoboTwin 2.0-Plus features a dual-arm bimanual system based on the Aloha-Agilex platform, with three cameras: a head-mounted third-person view and two wrist-mounted cameras, each capturing images at $320\times240$ resolution. These differences in embodiment, sensing configuration, and action space make the two benchmarks complementary. LIBERO-Plus primarily evaluates single-arm dexterity under perturbations, whereas RoboTwin 2.0-Plus focuses on the robustness of bimanual coordination.

\subsection{Evaluation Methods}

We conduct comprehensive evaluation of VLAs and WAMs for which publicly available checkpoints exist. On RoboTwin 2.0-Plus, We evaluate X-VLA~\citep{zheng2026xvla}, MOTUS~\citep{bi2025motus} and LingBot-VA~\citep{li2026causal} using their released checkpoints finetuned on the original RoboTwin 2.0 dataset. Due to the known performance gap between the JAX and PyTorch implementations of the $\pi$-series models---also reported in \citet{bi2025motus} and \citet{li2026causal}---and the absence of a JAX version of $\pi_{0.5}$, we finetuned the JAX implementation of $\pi_{0.5}$ on the RoboTwin 2.0 dataset. The model is finetuned from the pretrained $\pi_{0.5}$ checkpoint on the full 27.5k RoboTwin 2.0 training data for 60k gradient steps using the openpi framework's recommended configuration: AdamW   optimizer ($\beta_1{=}0.9$, $\beta_2{=}0.95$, gradient clipping at 1.0), cosine decay learning rate schedule (peak $2.5{\times}10^{-5}$, decaying to $2.5{\times}10^{-6}$), batch size 64, with delta joint actions. More model checkpoints are publicly available for LIBERO, and some of them have been evaluated on LIBERO-Plus. We report the results of a diverse set of VLAs and WAMs, including classic VLAs from the $\pi_0$-series~\citep{black2025pi0, pi05} (JAX version); approaches that incorporate world modeling as auxiliary tasks, such as VLA-JEPA~\citep{sun2026vla}; and recent WAMs including GE-Act~\citep{liao2025genie} and Cosmos-Policy~\citep{kim2026cosmos}. A detailed comparison of these models is presented in \Cref{app:model_comp}. 



Although DreamZero~\citep{ye2026dreamzero} is included in our WAM taxonomy (\Cref{tab:wam_compare}), we exclude it from the benchmark evaluation for three reasons. First, its released checkpoint is trained on a proprietary cross-embodiment dataset and cannot be directly applied to LIBERO or RoboTwin~2.0 without re-training. Second, re-training is prohibitively expensive, as the model is built on the Wan2.1-14B video generation backbone---the largest among all WAMs surveyed---and its autoregressive interleaved training procedure demands substantial GPU resources. Third, the inference pipeline requires a warm-up phase exceeding 15 minutes, making benchmark-scale evaluation across thousands of rollout episodes impractical. The checkpoint of GigaWorld-Policy is also not yet publicly available; we therefore include it in the architectural comparison but exclude it from the quantitative evaluation. Fast-WAM~\citep{yuan2026fast}, which was previously unreleased, has since become available and is included in our evaluation: we use the \texttt{robotwin\_uncond\_3cam\_384} checkpoint for RoboTwin 2.0-Plus (finetuned on the standard 27.5k clean + domain-randomized demonstrations) and a LIBERO checkpoint trained on clean demonstrations only. This training-data asymmetry between the two Fast-WAM checkpoints is intentional and revisited in \textbf{RQ 2}, as it offers a useful natural experiment on the role of task-specific data diversity.

\subsection{Results}

We aim to address the following research questions through our experiments and analysis: 
\begin{itemize}
    \item \textbf{RQ 1.} Are WAM-based policies robust to perturbations? 

    \item 
\textbf{RQ 2.} Is the performance advantage of WAMs consistent across different perturbation types? 

    \item 
\textbf{RQ 3.} How to explain the performance differences between VLAs and WAMs? 

\item \textbf{RQ 4.} What are the runtime characteristics of WAMs and how do they compare to those of VLAs?  
\end{itemize}

Our key findings addressing the aforementioned questions are summarized below.

\begin{table}[t]
\centering
\caption{\textbf{RoboTwin 2.0-Plus evaluation results}. The \textit{Original} column reports the success rate of each method evaluated in the original RoboTwin 2.0 \textit{Easy} setting. For a fair comparison, all methods use a single unified model across all tasks. The best result in each column is highlighted in \textbf{bold}, while second-best result is \underline{underlined}. \textbf{Lang.}: Language. \textbf{BG}: Background.}
\label{tab:robotwin_plus}
\begin{tabular}{@{}L{2.5cm} C{1.0cm} C{1.0cm} C{1.0cm} C{1.0cm} C{1.0cm} C{1.0cm} C{1.0cm} C{1.0cm} C{1.0cm}@{}}
\toprule
Model & Original & Camera & Robot & Lang. & Light & BG & Noise & Layout & Total \\ \midrule

\multicolumn{10}{c}{\textit{VLAs}} \\

\arrayrulecolor{lightgray}\midrule

$\pi_{0.5}$ & 78.4 & \textbf{45.6} & 27.6 & 74.4 & 49.6 & 71.7 & 64.9 & 56.8 & 58.6 \\
X-VLA & 65.6 & 23.2 & \underline{65.2} & 64.4 & 63.1 & 58.6  & 49.7 & 34.8 & 53.1 \\

\arrayrulecolor{black}\midrule

\multicolumn{10}{c}{\textit{VLA + WM}} \\
\arrayrulecolor{lightgray}\midrule

MOTUS & 87.0 & 21.6 & \textbf{85.0} & 83.2 & 84.6 & 84.4 & 43.1 & 82.8 & 71.5 \\

\arrayrulecolor{black}\midrule

\multicolumn{10}{c}{\textit{WAMs}} \\

\arrayrulecolor{lightgray}\midrule

LingBot-VA & \textbf{92.1} & 28.9  & 36.2 & \textbf{87.3}  & \textbf{89.0} & \textbf{91.3}  & \textbf{80.9} & \textbf{87.9} & \textbf{74.2}  \\
Fast-WAM & \underline{91.2} & \underline{30.4} & 53.2 & \underline{86.7} & \underline{88.8} & \underline{90.0} & \underline{76.4} & \underline{83.2} & \underline{72.7} \\
\arrayrulecolor{black}\bottomrule
\end{tabular}
\end{table}

\begin{figure}
    \centering
    \includegraphics[width=0.85\linewidth]{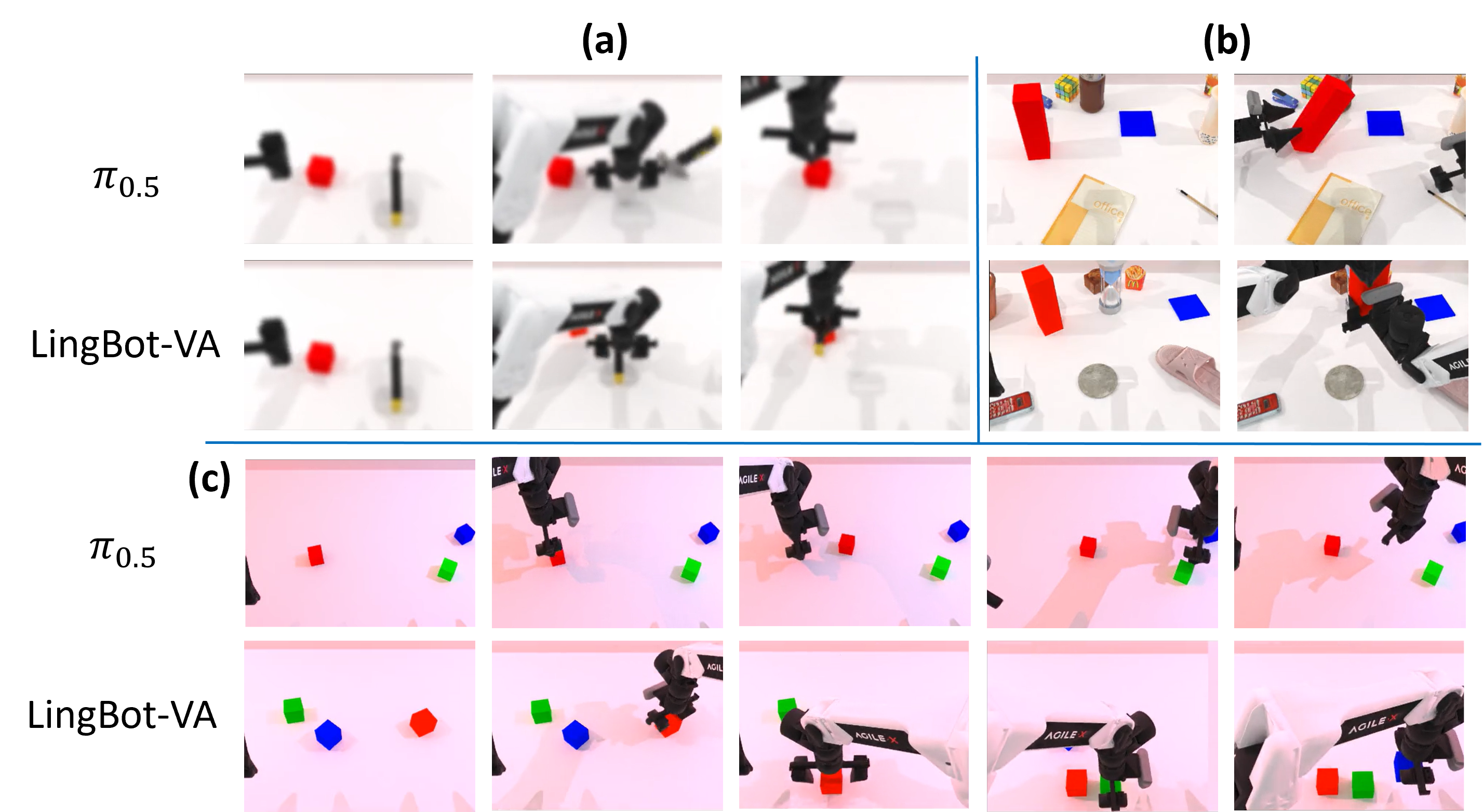}
    \caption{\textbf{Case studies on RoboTwin 2.0-Plus.} We present three representative cases comparing $\pi_{0.5}$ and LingBot-VA. Keyframes are shown sequentially from left to right.
    \textbf{(a)} \textit{Task:} beat block with hammer. \textit{Perturbation:} noise (N3). \textit{Result:} $\pi_{0.5}$ collides with the hammer and fails to complete the task.
    \textbf{(b)} \textit{Task:} handover block. \textit{Perturbation:} layout. \textit{Result:} $\pi_{0.5}$ collides with the red block during approach, leading to failure.
    \textbf{(c)} \textit{Task:} rank RGB blocks. \textit{Perturbation:} lighting (L1–L4 mix). \textit{Result:} $\pi_{0.5}$ fails to grasp the first red block due to misalignment and does not recover. LingBot-VA success in all three cases shown.
}
    \label{fig:robotwin_case}
\end{figure}

\textbf{RQ 1.} Are WAM-based policies robust to perturbations?

Recent WAM studies have reported superior performance over VLAs in both simulation and real robot experiments~\cite{kim2026cosmos, bi2025motus, li2026causal}. In particular, WAMs demonstrate stronger scene and task generalization, largely attributed to the spatiotemporal priors inherited from the video pre-training. We adopt RoboTwin 2.0-Plus and LIBERO-Plus described in prior session to rigorously evaluate the robustness of policy models under diverse perturbation conditions. As shown in \Cref{tab:robotwin_plus}, LinBot-VA achieves the SOTA success rate of 92.1\% on the original RoboTwin 2.0 benchmark and consistently exhibits great robustness to various perturbations. It ranks first in five out of seven perturbation categories and achieves an overall success rate of 74.2\%, which is significantly higher than 58.6\% by $\pi_{0.5}$ and 72.7\% of the second-best method Fast-WAM. Several representative cases are illustrated in \Cref{fig:robotwin_case}. While $\pi_{0.5}$ struggles with visual perturbations such as noise, light and cluttered layout, LingBot-VA performs effectively under these conditions, demonstrating greater robustness.
The results of LIBERO-Plus further corroborate the robustness advantage of WAMs on a second benchmark. Cosmos-Policy achieves an impressive 98.5\% success rate on the original LIBERO benchmark and maintains an 82.2\% overall success rate with all perturbations introduced in LIBERO-Plus. GE-Act follows closely, attaining an 80.3\% success rate on LIBERO-Plus. Despite the high sensitivity of most VLAs to perturbations, $\pi_{0.5}$ delivers strong and consistent performance, achieving the highest overall success rate of 85.7\%. 

Furthermore, Fast-WAM---the only WAM in our evaluation that requires no embodied pre-training---achieves an overall success rate of 72.7\% on RoboTwin 2.0-Plus (\Cref{tab:robotwin_plus}), placing it second only to LingBot-VA and ahead of MOTUS. This indicates that the spatiotemporal priors of the video backbone alone---without any large-scale embodied pre-training---can yield WAM-level robustness, provided the task-specific training data is sufficiently diverse, a condition we examine more carefully in \textbf{RQ 2}.

Notably, several models occupy a middle ground between pure VLAs and WAMs, and their behavior sheds light on how video priors are integrated, not merely whether they are present. MOTUS incorporates a pretrained video generation backbone (Wan2.2-5B) but routes action generation through a separate VLM expert rather than the video model itself. On RoboTwin 2.0-Plus as shown in \Cref{tab:robotwin_plus}, MOTUS achieves the highest robot initial-state robustness (85.0\%) and the third-best overall score (71.5\%), suggesting that incorporation of the additional video backbone and jointly training with dynamic prediction enhances the robustness of the learned policy. VLA-JEPA, on the other hand, augments a standard VLM backbone (Qwen3-VL-2B) with a future-state prediction objective trained on human videos, effectively injecting video-like temporal priors into the VLA framework. Its competitive performance on LIBERO-Plus (77.9\% in total) indicates that even partial incorporation of spatiotemporal learning can meaningfully improve robustness, though it still falls short of dedicated WAMs that natively operate in the video latent space.



\textbf{RQ 2.} Is the performance advantage of WAMs consistent across different perturbation types?

\begin{figure}[h]
    \centering
    \includegraphics[width=\linewidth]{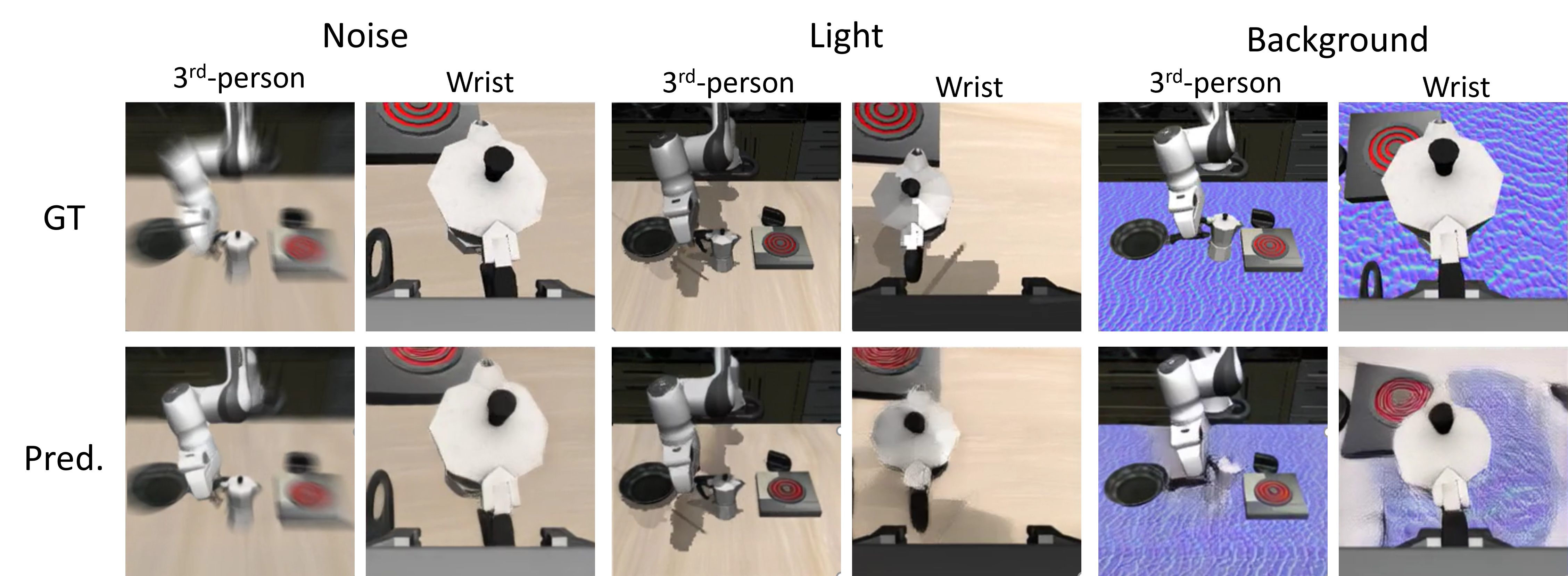}
    \caption{\textbf{Illustrations of future images predicted by Cosmos-policy.} Ground-truth and predicted images under three types of perturbations (noise, lighting, and background variations) in LIBERO-Plus are shown. GT: the ground truth. Pred.: the predictions generated by Cosmos-policy.}
    \label{fig:cosmos_pred}
\end{figure}

The behavior of WAMs varies across different types of perturbations. As shown in \Cref{tab:robotwin_plus}, LingBot-VA consistently demonstrates strong robustness to visually based perturbations. In particular, it maintains strong performance under light (89.0\%), noise (80.9\%), and layout (87.9\%) perturbations, while facing challenges in scenarios involving variations in camera viewpoint and the robot's initial state. A similar trend is observed in the LIBERO-Plus results shown in \Cref{tab:libero_plus_total}, where Cosmos-Policy and GE-Act demonstrate high robustness in light, noise and layout perturbations. Visualizations of predictions by Cosmos-Predict,  shown in \Cref{fig:cosmos_pred}, indicate that the future images predicted by Cosmos-policy remain highly accurate under varying noise and light conditions. Notably, when the input image is corrupted by noise, Cosmos-policy is able to effectively denoise the moving robotic arm in its predictions. This capacity is likely attributable to its pre-training on web-scale videos containing dynamic objects. However, the predictions exhibit severe spatial distortions and inconsistent color schemes under certain background perturbation---probably due to the uncommon background patterns. This degradation in the quality of predicted future images will likely lead to inaccurate action generation.

Fast-WAM further provides a particularly clean natural experiment on the role of task-specific training-data diversity. On RoboTwin 2.0-Plus, where the released checkpoint was finetuned on the standard 27.5k clean + domain-randomized demonstrations, Fast-WAM is highly robust: 91.2\% on the original benchmark and 72.7\% averaged across perturbations (\Cref{tab:robotwin_plus})---a drop of only $\sim$18 points. On LIBERO-Plus, where the same architecture was trained on clean demonstrations only, the model achieves 97.6\% on the original benchmark but degrades sharply under nearly every perturbation type (camera 16.4\%, background 53.7\%, noise 37.7\%; \Cref{tab:libero_plus_total}), corresponding to an overall $\sim$46-point drop. Since the architecture and video backbone are identical across the two checkpoints, the contrast isolates the contribution of training-data diversity: the spatiotemporal priors of the video backbone are necessary but not sufficient for robustness, and task-specific training-data diversity remains a critical lever even when the backbone is video-pretrained. Comparing across architectures, this also suggests that Fast-WAM---which jointly denoises state and action without explicitly conditioning the latter on a predicted future state---may be more dependent on training-data diversity to acquire generalization than IDM-style WAMs such as LingBot-VA, where the explicit state-to-action causal coupling provides an additional architectural prior for robustness. Overall, this is consistent with our broader finding that robustness emerges from the \emph{integration} of dynamic priors with diverse training data rather than from the presence of either alone.

\begin{table}[t]
\centering
\caption{\textbf{LIBERO-Plus evaluation results}. The \textit{Original} column denotes results on the standard LIBERO benchmark. The results of $\pi_0$, $\pi_0\text{-FAST}$, Openvla-OFT-m, UniVLA, RIPT-VLA are sourced from \citet{fei25libero-plus}. The results reported for HoloBrain0-GD~\citep{lin2026holobrain} and ABot-M0~\citep{yang2026abot} are taken from their respective papers. The remaining models are evaluated using their official repositories and publicly released checkpoints. Specifically, $\pi_0 (\text{rerun})$ and $\pi_{0.5}$ are evaluated using their JAX implementations. The best result in each column is highlighted in \textbf{bold}, while second-best result is \underline{underlined}. \textbf{Lang.}: Language. \textbf{BG}: Background.}


\label{tab:libero_plus_total}
\begin{tabular}{@{}L{3.0cm} C{1.0cm} C{1.0cm} C{1.0cm} C{1.0cm} C{1.0cm} C{1.0cm} C{1.0cm} C{1.0cm} C{1.0cm}}
\toprule
Model & Original & Camera & Robot & Lang. & Light & BG & Noise & Layout & Total \\ \midrule

\multicolumn{10}{c}{\textit{VLAs}} \\

\arrayrulecolor{lightgray}\midrule

$\pi_0$ & 94.2 & 13.8 & 6.0 & 58.8 & 85.0 & 81.4 & 79.0 & 68.9 & 53.6 \\
$\pi_0 (\text{rerun})$ & 91.3 & 61.0 & 40.8 & 63.5 & 89.3 & 84.1 & 80.1 & 76.4 & 69.4 \\
$\pi_0$-FAST & 85.5 & 65.1 & 21.6 & 61.0 & 73.2 & 73.2 & 74.4 & 68.8 & 61.6 \\
$\pi_{0.5}$ & 96.9 & \underline{75.4} & \underline{77.5} & 85.6 & \textbf{96.9} & \underline{94.6} & 89.7 & \textbf{85.7} & \textbf{85.7} \\
Openvla-OFT\_m & 97.6 & 55.6 & 21.7 & 81.0 & 92.7 & 91.0 & 78.6 & 68.7 & 67.9 \\
UniVLA & 95.2 & 1.8 & 46.2 & 69.6 & 69.0 & 81.0 & 21.2 & 31.9 & 42.9 \\
RIPT-VLA & 97.5 & 55.2 & 31.2 & 77.6 & 88.4 & 91.6 & 73.5 & 74.2 & 68.4 \\
X-VLA & 98.1 & 23.4 & \textbf{89.7} & 75.7 & 88.2 & \textbf{96.0} & 62.7 & 71.8 & 71.4 \\
HoloBrain0-GD & 96.7 & 65.5 & 58.2 & 78.7 & 88.1 & 90.3 & 66.9 & 79.5 & 74.0 \\
ABot-M0 & \textbf{98.6} & 60.4 & 67.9 & \underline{86.4} & 96.2 & 91.6 & 86.4 & 82.6 & 80.5 \\ 

\arrayrulecolor{black}\midrule
\multicolumn{10}{c}{\textit{VLA + WM}} \\
\arrayrulecolor{lightgray}\midrule

VLA-JEPA & 97.2 & 64.2 & 67.7 & \textbf{88.1} & 91.8 & 93.4 & 65.8 & \underline{83.9} & 77.9 \\ 

\arrayrulecolor{black}\midrule
\multicolumn{10}{c}{\textit{WAMs}} \\
\arrayrulecolor{lightgray}\midrule

GE-Act & 94.4 & 60.7 & 77.0 & 77.4 & 95.8 & 86.0 & \underline{90.9} & 80.2 & 80.3 \\
Cosmos-Policy & \underline{98.5} & \textbf{75.8} & 63.3 & 81.7 & \underline{96.5} & 88.9 & \textbf{92.7} & 82.2 & \underline{82.2} \\
Fast-WAM & 97.6 & 16.4 & 44.5 & 68.9 & 78.2 & 53.7 & 37.7 & 60.7 & 51.5 \\
\arrayrulecolor{black}\bottomrule
\end{tabular}
\end{table}

\textbf{RQ 3.} How to explain the performance differences between VLAs and WAMs?

As discussed in \Cref{sec:diff_wam}, WAMs differ from VLAs in both backbone selection and the subsequent training strategy for robot policy generation. Unlike VLAs, which typically leverage VLM backbones, WAMs build upon video diffusion models pretrained on web-scale video generation tasks. Exposure to diverse and temporally rich video data in video backbone enables WAMs to capture fine-grained visual dynamics and temporal dependencies without requiring extensive training for policy generation. As shown in \Cref{tab:training_data}, WAMs such as Cosmos-policy require only lightweight finetuning on downstream tasks while maintaining strong robustness under various visual perturbations. In contrast, the backbone models of VLAs are primarily trained on static image-text data and lack explicit priors over visual dynamics, which can limit their ability to adapt to new scenes and tasks. To achieve robustness comparable to WAMs, VLAs typically require additional training on diverse dynamic visual data. For instance, $\pi_{0.5}$ achieves strong cross-scene generalization and robustness to visual perturbations by incorporating diverse robotic data and web data into its training pipeline. Similarly, hybrid methods such as VLA-JEPA and MOTUS with strong robustness also incorporate human video data in their training pipeline. Overall, the spatiotemporal priors inherited from video generation backbones substantially simplifies the training process of WAMs and reduce their reliance on diverse visual data. This advantage is particularly valuable in settings where data availability is limited.


\begin{table}[h]
\caption{\textbf{Runtime Speed Comparison.} The inference times of all models are evaluated using the same device. LingBot-VA(RW): real-world setting with three denoising steps for state and five for action. LingBot-VA(RT): RoboTwin 2.0 setting with 25 denoising steps for state and 50 for action. }
\label{tab:speed}
\centering
\begin{tabularx}{\linewidth}{
>{\raggedright\arraybackslash\hsize=1.8\hsize}X
*{9}{>{\centering\arraybackslash\hsize=0.7\hsize}X}
}
\toprule
 & $\pi_{0.5}$ & X-VLA & Fast-WAM$^\ast$ & GE-Act & Cosmos-Policy & LingBot-VA(RW) & MOTUS & LingBot-VA(RT) \\
\midrule
Action chunk size & 50 & 30 & 32 & 36 & 16 & 32 & 16 & 32 \\
Inference $t$ (wall clock) & 63 ms & 195 ms & 190 ms & 300 ms & 390 ms & 480 ms & 1175 ms & 5230 ms \\
Inference $t$ (w.r.t.\ $\pi_{0.5}$) & 1.0 & 3.1 $\times$ & 3.0 $\times$ & 4.8 $\times$ & 6.2 $\times$ & 7.6 $\times$ & 18.6 $\times$ & 83.0 $\times$ \\
\bottomrule
\end{tabularx}
{\footnotesize $^\ast$Fast-WAM latency is reported by \citet{yuan2026fast} on their hardware and is not re-measured on our device.}
\end{table}

\textbf{RQ 4.} What are the runtime characteristics of WAMs and how do they compare to those of VLAs?

A major limitation of WAMs is their slower runtime compared to VLAs, primarily due to the computational overhead of the future-state diffusion process. As summarized in \Cref{tab:speed}, the inference cost of WAMs varies considerably depending on architecture choices. Among VLAs, $\pi_{0.5}$ achieves the fastest inference at 63\,ms per chunk, while X-VLA takes 195\,ms per inference. In contrast, MOTUS and WAMs are consistently slower, requiring at least 4.8$\times$ runtime per inference compared to $\pi_{0.5}$.
A critical factor that largely determines the speed of WAMs is the number of denoising steps for state and action, with state denoising typically dominating the runtime. For example, GE-Act requires only one denoising step for state and ten for action, whereas Cosmos-Policy performs five steps for both. MOTUS, in contrast, uses ten denoising steps for both state and action. LingBot-VA adopts different configurations depending on the setting: for real-world deployment, it uses three denoising steps for state and five for action, while for RoboTwin 2.0, it employs 25 and 50 steps, respectively, which explains its significantly longer inference time (5.2 seconds per inference). Runtime speed of WAMs is also affected by other factors, including backbone size (2B--5.3B parameters), generation strategy (joint denoising vs.\ separate action decoding), and whether autoregressive generation is employed. By omitting visual state generation at test time, Fast-WAM achieves the lowest inference latency among WAMs in our comparison (190 ms; \Cref{tab:speed}), but is still 3.0$\times$ slower than $\pi_{0.5}$. Further efforts are still needed to improve the inference efficiency of WAMs.

\section{Conclusion}

We presented a systematic comparative study of world action models (WAMs) and vision-language-action models (VLAs), evaluating their robustness under diverse visual and language perturbations across two complementary benchmarks: RoboTwin 2.0-Plus and LIBERO-Plus. Our findings reveal that WAMs, by leveraging spatiotemporal priors from video generation backbones, consistently demonstrate strong robustness to noise, lighting, and layout perturbations---a pattern that holds across both single-arm (LIBERO) and bimanual (RoboTwin) settings. However, camera viewpoint and robot initial-state perturbations remain challenging for WAMs, indicating that video priors offer limited benefit when the geometric configuration of the scene is altered. Among VLAs, $\pi_{0.5}$ achieves competitive or superior robustness by incorporating diverse robotic and web data during training, suggesting that data diversity can compensate for the absence of explicit world modeling. Hybrid approaches such as MOTUS and VLA-JEPA, which partially integrate video-based temporal learning, exhibit robustness profiles that fall between pure VLAs and WAMs, highlighting that the manner in which video priors are incorporated matters as much as their presence.

Our analysis also reveals that inference speed remains a major challenge for WAMs, primarily driven by the visual state denoising process. While faster methods such as GE-Act requires only a single denoising step, the slowest model, LingBot-VA, requires up to 25 steps on the RoboTwin dataset. Compared to $\pi_{0.5}$, the evaluated WAMs are at least 4.8 times slower per inference. This underscores a key practical challenge for WAMs and highlights the need for further research to improve their inference efficiency, enabling deployment in scenarios that require rapid response time or real-time interaction with dynamic environments. Recent approaches, such as Fast-WAM, have sought to accelerate inference by co-training visual state and action prediction and omitting state generation at test time. In our evaluation, Fast-WAM achieves competitive robustness on RoboTwin 2.0-Plus (72.7\% overall, second only to LingBot-VA) without any embodied pre-training, demonstrating the data efficiency of the WAM paradigm. However, when restricted to clean training demonstrations on LIBERO, Fast-WAM's robustness collapses (51.5\% overall vs.\ 97.6\% on the original benchmark), reinforcing that the WAM video prior is necessary but not sufficient---task-specific training-data diversity remains essential. A substantial latency gap also remains relative to $\pi_{0.5}$ ($\sim$3$\times$) in real-world deployment, indicating room for further optimization.



\newpage

\bibliographystyle{plainnat}
\bibliography{references}

@article{lin2026holobrain,
  title={HoloBrain-0 Technical Report},
  author={Lin, Xuewu and Lin, Tianwei and Du, Yun and Xie, Hongyu and Jin, Yiwei and Li, Jiawei and Wu, Shijie and Wang, Qingze and Li, Mengdi and Zhao, Mengao and others},
  journal={arXiv preprint arXiv:2602.12062},
  year={2026}
}

@inproceedings{bar2025navigation,
  title={Navigation world models},
  author={Bar, Amir and Zhou, Gaoyue and Tran, Danny and Darrell, Trevor and LeCun, Yann},
  booktitle={Proceedings of the Computer Vision and Pattern Recognition Conference},
  pages={15791--15801},
  year={2025}
}

@article{yang2025mindjourney,
  title={MindJourney: Test-Time Scaling with World Models for Spatial Reasoning},
  author={Yang, Yuncong and Liu, Jiageng and Zhang, Zheyuan and Zhou, Siyuan and Tan, Reuben and Yang, Jianwei and Du, Yilun and Gan, Chuang},
  journal={arXiv preprint arXiv:2507.12508},
  year={2025}
}

@article{shen2026efficient,
  title={An Efficient and Multi-Modal Navigation System with One-Step World Model},
  author={Shen, Wangtian and Meng, Ziyang and Ma, Jinming and Zhou, Mingliang and Xiang, Diyun},
  journal={arXiv preprint arXiv:2601.12277},
  year={2026}
}

@article{dong2025unified,
  title={Unified world models: Memory-augmented planning and foresight for visual navigation},
  author={Dong, Yifei and Wu, Fengyi and Chen, Guangyu and Cheng, Zhi-Qi and Hu, Qiyu and Zhou, Yuxuan and Sun, Jingdong and He, Jun-Yan and Dai, Qi and Hauptmann, Alexander G},
  journal={arXiv preprint arXiv:2510.08713},
  year={2025}
}

@article{kouvaritakis2016model,
  title={Model predictive control},
  author={Kouvaritakis, Basil and Cannon, Mark},
  journal={Switzerland: Springer International Publishing},
  volume={38},
  number={13-56},
  pages={7},
  year={2016},
  publisher={Springer}
}

@article{richens2025general,
  title={General agents contain world models},
  author={Richens, Jonathan and Abel, David and Bellot, Alexis and Everitt, Tom},
  journal={arXiv preprint arXiv:2506.01622},
  year={2025}
}

@inproceedings{Yuanmin_2025_CVPR,
  author={Yuanmin Tang and Jing Yu and Keke Gai and Jiamin Zhuang and Gang Xiong and Gaopeng Gou and Qi Wu},
  title={Missing Target-Relevant Information Prediction with World Model for Accurate Zero-Shot Composed Image Retrieval},
  year={2025},
  booktitle={CVPR}
}

@inproceedings{Hassan_2025_CVPR,
  author={Mariam Hassan and Sebastian Stapf and Ahmad Rahimi and Pedro M. B. Rezende and Yasaman Haghighi and David Brüggemann and Isinsu Katircioglu and Lin Zhang and Xiaoran Chen and Suman Saha and Marco Cannici and Elie Aljalbout and Botao Ye and Xi Wang and Aram Davtyan and Mathieu Salzmann and Davide Scaramuzza and Marc Pollefeys and Paolo Favaro and Alexandre Alahi},
  title={GEM: A Generalizable Ego-Vision Multimodal World Model for Fine-Grained Ego-Motion, Object Dynamics, and Scene Composition Control},
  year={2025},
  booktitle={CVPR}
}

@inproceedings{Zhao_2025_CVPR,
  author={Guosheng Zhao and Chaojun Ni and Xiaofeng Wang and Zheng Zhu and Xueyang Zhang and Yida Wang and Guan Huang and Xinze Chen and Boyuan Wang and Youyi Zhang and Wenjun Mei and Xingang Wang},
  title={DriveDreamer4D: World Models Are Effective Data Machines for 4D Driving Scene Representation},
  year={2025},
  booktitle={CVPR}
}

@inproceedings{Yue_2025_CVPR,
  author={Yang Yue and Yulin Wang and Haojun Jiang and Pan Liu and Shiji Song and Gao Huang},
  title={EchoWorld: Learning Motion-Aware World Models for Echocardiography Probe Guidance},
  year={2025},
  booktitle={CVPR}
}

@inproceedings{yang2025medical,
  title={Medical world model},
  author={Yang, Yijun and Wang, Zhao-Yang and Liu, Qiuping and Sun, Shuwen and Wang, Kang and Chellappa, Rama and Zhou, Zongwei and Yuille, Alan and Zhu, Lei and Zhang, Yu-Dong and others},
  year={2025},
  booktitle={ICCV}
}

@inproceedings{zheng2025learning,
  title={Learning Counterfactually Decoupled Attention for Open-World Model Attribution},
  author={Zheng, Yu and Gong, Boyang and Kong, Fanye and Duan, Yueqi and Yu, Bingyao and Zheng, Wenzhao and Chen, Lei and Lu, Jiwen and Zhou, Jie},
  year={2025},
  booktitle={ICCV}
}

@inproceedings{hao2025neural,
  title={Neural Motion Simulator Pushing the Limit of World Models in Reinforcement Learning},
  author={Hao, Chenjie and Lu, Weyl and Xu, Yifan and Chen, Yubei},
  year={2025},
  booktitle={CVPR}
}

@inproceedings{wang2025disentangled,
  title={Disentangled World Models: Learning to Transfer Semantic Knowledge from Distracting Videos for Reinforcement Learning},
  author={Wang, Qi and Zhang, Zhipeng and Xie, Baao and Jin, Xin and Wang, Yunbo and Wang, Shiyu and Zheng, Liaomo and Yang, Xiaokang and Zeng, Wenjun},
  year={2025},
  booktitle={ICCV}
}

@inproceedings{yao2025navmorph,
  title={NavMorph: A Self-Evolving World Model for Vision-and-Language Navigation in Continuous Environments},
  author={Yao, Xuan and Gao, Junyu and Xu, Changsheng},
  year={2025},
  booktitle={ICCV}
}

@inproceedings{zhen2025learning,
  title={Learning 4D Embodied World Models},
  author={Zhen, Haoyu and Sun, Qiao and Zhang, Hongxin and Li, Junyan and Zhou, Siyuan and Du, Yilun and Gan, Chuang},
  year={2025},
  booktitle={ICCV}
}

@inproceedings{goswami2025osvi,
  title={Osvi-wm: One-shot visual imitation for unseen tasks using world-model-guided trajectory generation},
  author={Goswami, Raktim Gautam and Krishnamurthy, Prashanth and LeCun, Yann and Khorrami, Farshad},
  year={2025},
  booktitle={NeurIPS}
}

@inproceedings{shang2025roboscape,
  title={RoboScape: Physics-informed Embodied World Model},
  author={Shang, Yu and Zhang, Xin and Tang, Yinzhou and Jin, Lei and Gao, Chen and Wu, Wei and Li, Yong},
  year={2025},
  booktitle={NeurIPS}
}

@article{sandage1988observational,
  title={Observational tests of world models},
  author={Sandage, Allan},
  journal={Annual Review of Astronomy and Astrophysics},
  volume={26},
  pages={561--630},
  year={1988}
}

@article{ha2018world,
  title={World models},
  author={Ha, David and Schmidhuber, J{\"u}rgen},
  journal={arXiv preprint arXiv:1803.10122},
  year={2018}
}

@book{craik1967nature,
  title={The nature of explanation},
  author={Craik, Kenneth James Williams},
  volume={445},
  year={1967},
  publisher={Cambrdige University Press}
}

@inproceedings{hafner2019_planet,
  title        = {Learning Latent Dynamics for Planning from Pixels},
  author       = {Hafner, Danijar and Lillicrap, Timothy and Fischer, Ian and Villegas, Ruben and Ha, David and Lee, Honglak and Davidson, James},
  booktitle    = {Proceedings of the 36th International Conference on Machine Learning (ICML)},
  year         = {2019},
  url          = {https://proceedings.mlr.press/v97/hafner19a.html}
}

@inproceedings{hafner2020_dreamer,
  title        = {Dream to Control: Learning Behaviors by Latent Imagination},
  author       = {Hafner, Danijar and Lillicrap, Timothy and Ba, Jimmy and Norouzi, Mohammad},
  booktitle    = {International Conference on Learning Representations (ICLR)},
  year         = {2020},
  url          = {https://openreview.net/forum?id=S1lOTC4tDS}
}

@article{lambert2020_objective_mismatch,
  title        = {Objective Mismatch in Model-based Reinforcement Learning},
  author       = {Lambert, Nathan and Amos, Brandon and Yadan, Omry and Calandra, Roberto},
  journal      = {arXiv preprint arXiv:2002.04523},
  year         = {2020},
  eprint       = {2002.04523},
  archivePrefix= {arXiv},
  primaryClass = {cs.LG},
  note         = {Published in L4DC 2020},
  url          = {https://arxiv.org/abs/2002.04523}
}

@article{littman2001_psr,
  title        = {Predictive Representations of State},
  author       = {Littman, Michael L. and Sutton, Richard S.},
  journal      = {Advances in Neural Information Processing Systems},
  volume       = {14},
  year         = {2001},
  url          = {https://papers.nips.cc/paper_files/paper/2001/hash/1e4d36177d71bbb3558e43af9577d70e-Abstract.html}
}

@article{finn2017_visualforesight,
  title        = {Deep Visual Foresight for Planning Robot Motion},
  author       = {Finn, Chelsea and Levine, Sergey},
  journal      = {arXiv preprint arXiv:1610.00696},
  year         = {2017},
  eprint       = {1610.00696},
  archivePrefix= {arXiv},
  primaryClass = {cs.LG},
  note         = {ICRA 2017},
  url          = {https://arxiv.org/abs/1610.00696}
}

@article{ha2018_worldmodels,
  title        = {World Models},
  author       = {Ha, David and Schmidhuber, J{\"u}rgen},
  journal      = {arXiv preprint arXiv:1803.10122},
  year         = {2018},
  eprint       = {1803.10122},
  archivePrefix= {arXiv},
  primaryClass = {cs.LG},
  url          = {https://arxiv.org/abs/1803.10122}
}

@article{chua2018_pets,
  title        = {Deep Reinforcement Learning in a Handful of Trials using Probabilistic Dynamics Models},
  author       = {Chua, Kurtland and Calandra, Roberto and McAllister, Rowan and Levine, Sergey},
  journal      = {arXiv preprint arXiv:1805.12114},
  year         = {2018},
  eprint       = {1805.12114},
  archivePrefix= {arXiv},
  primaryClass = {cs.LG},
  note         = {NeurIPS 2018 (PETS)},
  url          = {https://arxiv.org/abs/1805.12114}
}

@article{ebert2018_visualforesight,
  title        = {Visual Foresight: Model-Based Deep Reinforcement Learning for Vision-Based Robotic Control},
  author       = {Ebert, Frederik and Finn, Chelsea and Dasari, Sudeep and Xie, Annie and Lee, Alex and Levine, Sergey},
  journal      = {arXiv preprint arXiv:1812.00568},
  year         = {2018},
  eprint       = {1812.00568},
  archivePrefix= {arXiv},
  primaryClass = {cs.RO},
  url          = {https://arxiv.org/abs/1812.00568}
}

@article{janner2019_mbpo,
  title        = {When to Trust Your Model: Model-Based Policy Optimization},
  author       = {Janner, Michael and Fu, Justin and Zhang, Marvin and Levine, Sergey},
  journal      = {arXiv preprint arXiv:1906.08253},
  year         = {2019},
  eprint       = {1906.08253},
  archivePrefix= {arXiv},
  primaryClass = {cs.LG},
  note         = {NeurIPS 2019},
  url          = {https://arxiv.org/abs/1906.08253}
}

@inproceedings{hansen2022_tdmpc,
  title        = {Temporal Difference Learning for Model Predictive Control},
  author       = {Hansen, Nicklas and Wang, Xiaolong and Su, Hao},
  booktitle    = {Proceedings of the 39th International Conference on Machine Learning (ICML)},
  year         = {2022},
  url          = {https://proceedings.mlr.press/v162/hansen22a.html}
}

@article{hafner2023_dreamerv3,
  title        = {Mastering Diverse Domains through World Models},
  author       = {Hafner, Danijar and Pasukonis, Jurgis and Ba, Jimmy and Lillicrap, Timothy},
  journal      = {arXiv preprint arXiv:2301.04104},
  year         = {2023},
  eprint       = {2301.04104},
  archivePrefix= {arXiv},
  primaryClass = {cs.AI},
  note         = {DreamerV3 (arXiv)},
  url          = {https://arxiv.org/abs/2301.04104}
}

@article{hansen2024_tdmpc2,
  title        = {TD-MPC2: Scalable, Robust World Models for Continuous Control},
  author       = {Hansen, Nicklas and Su, Hao and Wang, Xiaolong},
  journal      = {arXiv preprint arXiv:2310.16828},
  year         = {2024},
  eprint       = {2310.16828},
  archivePrefix= {arXiv},
  primaryClass = {cs.LG},
  note         = {ICLR 2024},
  url          = {https://arxiv.org/abs/2310.16828}
}

@article{soh2026_action_hallucination,
  title        = {Action Hallucination in Generative Visual-Language-Action Models},
  author       = {Soh, Harold and Lim, Eugene},
  journal      = {arXiv preprint arXiv:2602.06339},
  year         = {2026},
  eprint       = {2602.06339},
  archivePrefix= {arXiv},
  primaryClass = {cs.RO},
  url          = {https://arxiv.org/abs/2602.06339}
}

@article{assran2025vjepa2,
  title   = {V-JEPA~2: Self-Supervised Video Models Enable Understanding, Prediction and Planning},
  author  = {Assran, Mahmoud and Bardes, Adrien and Fan, David and Garrido, Quentin and Howes, Russell and
             Komeili, Mojtaba and Muckley, Matthew and Rizvi, Ammar and Roberts, Claire and Sinha, Koustuv and
             Zholus, Artem and Arnaud, Sergio and Gejji, Abha and Martin, Ada and Robert Hogan, Francois and
             Dugas, Daniel and Bojanowski, Piotr and Khalidov, Vasil and Labatut, Patrick and Massa, Francisco and
             Szafraniec, Marc and Krishnakumar, Kapil and Li, Yong and Ma, Xiaodong and Chandar, Sarath and
             Meier, Franziska and LeCun, Yann and Rabbat, Michael and Ballas, Nicolas},
  journal = {arXiv preprint arXiv:2506.09985},
  year    = {2025},
  url     = {https://arxiv.org/abs/2506.09985}
}

@misc{nvidia2025_cosmos,
  title        = {NVIDIA Cosmos},
  author       = {NVIDIA},
  howpublished = {Product / documentation page},
  year         = {2025},
  url          = {https://www.nvidia.com/en-us/ai/cosmos/}
}

@article{ali2025_cosmospredict25,
  title   = {World Simulation with Video Foundation Models for Physical AI},
  author  = {NVIDIA and Arslan Ali and Junjie Bai and Maciej Bala and Yogesh Balaji and Aaron Blakeman and
             Tiffany Cai and Jiaxin Cao and Tianshi Cao and Elizabeth Cha and Yu-Wei Chao and Prithvijit Chattopadhyay and
             Mike Chen and Yongxin Chen and Yu Chen and Shuai Cheng and Yin Cui and Jenna Diamond and Yifan Ding and
             Jiaojiao Fan and Linxi Fan and Liang Feng and Francesco Ferroni and Sanja Fidler and Xiao Fu and Ruiyuan Gao and
             Yunhao Ge and Jinwei Gu and Aryaman Gupta and Siddharth Gururani and Imad El Hanafi and Ali Hassani and
             Zekun Hao and Jacob Huffman and Joel Jang and Pooya Jannaty and Jan Kautz and Grace Lam and Xuan Li and
             Zhaoshuo Li and Maosheng Liao and Chen-Hsuan Lin and Tsung-Yi Lin and Yen-Chen Lin and Huan Ling and
             Ming-Yu Liu and Xian Liu and Yifan Lu and Alice Luo and Qianli Ma and Hanzi Mao and Kaichun Mo and
             Seungjun Nah and Yashraj Narang and Abhijeet Panaskar and Lindsey Pavao and Trung Pham and Morteza Ramezanali and
             Fitsum Reda and Scott Reed and Xuanchi Ren and Haonan Shao and Yue Shen and Stella Shi and Shuran Song and
             Bartosz Stefaniak and Shangkun Sun and Shitao Tang and Sameena Tasmeen and Lyne Tchapmi and Wei-Cheng Tseng and
             Jibin Varghese and Andrew Z. Wang and Hao Wang and Haoxiang Wang and Heng Wang and Ting-Chun Wang and
             Fangyin Wei and Jiashu Xu and Dinghao Yang and Xiaodong Yang and Haotian Ye and Seonghyeon Ye and
             Xiaohui Zeng and Jing Zhang and Qinsheng Zhang and Kaiwen Zheng and Andrew Zhu and Yuke Zhu},
  journal = {arXiv preprint arXiv:2511.00062},
  year    = {2025},
  url     = {https://arxiv.org/abs/2511.00062}
}

@inproceedings{hu2025video,
title={Video Prediction Policy: A Generalist Robot Policy with Predictive Visual Representations},
author={Yucheng Hu and Yanjiang Guo and Pengchao Wang and Xiaoyu Chen and Yen-Jen Wang and Jianke Zhang and Koushil Sreenath and Chaochao Lu and Jianyu Chen},
booktitle={Forty-second International Conference on Machine Learning},
year={2025},
url={https://openreview.net/forum?id=c0dhw1du33}
}

@article{pai2025mimicvideo,
  author    = {Jonas Pai and Liam Achenbach and Victoriano Montesinos and Benedek Forrai and Oier Mees and Elvis Nava},
  title     = {mimic-video: Video-Action Models for Generalizable Robot Control Beyond VLAs},
  journal   = {arXiv preprint 2512.15692},
  year      = {2025},
}

@InProceedings{Chen_2025_ICCV,
    author    = {Chen, Yi and Ge, Yuying and Tang, Weiliang and Li, Yizhuo and Ge, Yixiao and Ding, Mingyu and Shan, Ying and Liu, Xihui},
    title     = {Moto: Latent Motion Token as the Bridging Language for Learning Robot Manipulation from Videos},
    booktitle = {Proceedings of the IEEE/CVF International Conference on Computer Vision (ICCV)},
    month     = {October},
    year      = {2025},
    pages     = {19752-19763}
}

@article{cen2025worldvla,
  title={Worldvla: Towards autoregressive action world model},
  author={Cen, Jun and Yu, Chaohui and Yuan, Hangjie and Jiang, Yuming and Huang, Siteng and Guo, Jiayan and Li, Xin and Song, Yibing and Luo, Hao and Wang, Fan and others},
  journal={arXiv preprint arXiv:2506.21539},
  year={2025}
}

@inproceedings{zhang2025dreamvla,
title={Dream{VLA}: A Vision-Language-Action Model Dreamed with Comprehensive World Knowledge},
author={Wenyao Zhang and Hongsi Liu and Zekun Qi and Yunnan Wang and XinQiang Yu and Jiazhao Zhang and Runpei Dong and Jiawei He and He Wang and Zhizheng Zhang and Li Yi and Wenjun Zeng and Xin Jin},
booktitle={The Thirty-ninth Annual Conference on Neural Information Processing Systems},
year={2025},
url={https://openreview.net/forum?id=PK07eretkF}
}

@article{sun2026vla,
  title={VLA-JEPA: Enhancing Vision-Language-Action Model with Latent World Model},
  author={Sun, Jingwen and Zhang, Wenyao and Qi, Zekun and Ren, Shaojie and Liu, Zezhi and Zhu, Hanxin and Sun, Guangzhong and Jin, Xin and Chen, Zhibo},
  journal={arXiv preprint arXiv:2602.10098},
  year={2026}
}

@article{zhu2025unified,
  title={Unified world models: Coupling video and action diffusion for pretraining on large robotic datasets},
  author={Zhu, Chuning and Yu, Raymond and Feng, Siyuan and Burchfiel, Benjamin and Shah, Paarth and Gupta, Abhishek},
  journal={arXiv preprint arXiv:2504.02792},
  year={2025}
}

@article{li2025unified,
  title={Unified video action model},
  author={Li, Shuang and Gao, Yihuai and Sadigh, Dorsa and Song, Shuran},
  journal={arXiv preprint arXiv:2503.00200},
  year={2025}
}

@inproceedings{driess2023palm,
  title={PaLM-E: an embodied multimodal language model},
  author={Driess, Danny and Xia, Fei and Sajjadi, Mehdi SM and Lynch, Corey and Chowdhery, Aakanksha and Ichter, Brian and Wahid, Ayzaan and Tompson, Jonathan and Vuong, Quan and Yu, Tianhe and others},
  booktitle={Proceedings of the 40th International Conference on Machine Learning},
  pages={8469--8488},
  year={2023}
}

@inproceedings{zhao2025cotvla,
  title={CoT-VLA: Visual Chain-of-Thought Reasoning for Vision-Language-Action Models},
  author={Zhao, Qingqing and Lu, Yao and Kim, Moo Jin and others},
  booktitle={Proceedings of the IEEE/CVF Conference on Computer Vision and Pattern Recognition (CVPR)},
  year={2025}
}

@inproceedings{zhou2025chatvla2,
  title={ChatVLA-2: Vision-Language-Action Model with Open-World Reasoning},
  author={Zhou, Zhongyi and Zhu, Yichen and Liu, Xiaoyu and others},
  booktitle={Advances in Neural Information Processing Systems (NeurIPS)},
  year={2025}
}

@inproceedings{li2026simplevlarl,
  title={SimpleVLA-RL: Scaling VLA Training via Reinforcement Learning},
  author={Li, Haozhan and Zuo, Yuxin and Yu, Jiale and others},
  booktitle={International Conference on Learning Representations (ICLR)},
  year={2026}
}

@InProceedings{openvla2025,
  title = 	 {OpenVLA: An Open-Source Vision-Language-Action Model},
  author =       {Kim, Moo Jin and Pertsch, Karl and Karamcheti, Siddharth and Xiao, Ted and Balakrishna, Ashwin and Nair, Suraj and Rafailov, Rafael and Foster, Ethan P and Sanketi, Pannag R and Vuong, Quan and Kollar, Thomas and Burchfiel, Benjamin and Tedrake, Russ and Sadigh, Dorsa and Levine, Sergey and Liang, Percy and Finn, Chelsea},
  booktitle = 	 {CoRL},
  year = 	 {2025}
  }

@InProceedings{pi05,
  title = 	 {$\pi_{0.5}$: a Vision-Language-Action Model with Open-World Generalization},
  author =       {Black, Kevin and Brown, Noah and Darpinian, James and Dhabalia, Karan and Driess, Danny and Esmail, Adnan and Equi, Michael Robert and Finn, Chelsea and Fusai, Niccolo and Galliker, Manuel Y. and Ghosh, Dibya and Groom, Lachy and Hausman, Karol and ichter, brian and Jakubczak, Szymon and Jones, Tim and Ke, Liyiming and LeBlanc, Devin and Levine, Sergey and Li-Bell, Adrian and Mothukuri, Mohith and Nair, Suraj and Pertsch, Karl and Ren, Allen Z. and Shi, Lucy Xiaoyang and Smith, Laura and Springenberg, Jost Tobias and Stachowicz, Kyle and Tanner, James and Vuong, Quan and Walke, Homer and Walling, Anna and Wang, Haohuan and Yu, Lili and Zhilinsky, Ury},
  booktitle = 	 {Proceedings of The 9th Conference on Robot Learning},
  year = 	 {2025}
  }

@article{elbanhawi2014sampling,
  title={Sampling-based robot motion planning: A review},
  author={Elbanhawi, Mohamed and Simic, Milan},
  journal={Ieee access},
  volume={2},
  pages={56--77},
  year={2014}
}

@inproceedings{zheng2026xvla,
title={X-{VLA}: Soft-Prompted Transformer as Scalable Cross-Embodiment Vision-Language-Action Model},
author={Jinliang Zheng and Jianxiong Li and Zhihao Wang and Dongxiu Liu and Xirui Kang and Yuchun Feng and Yinan Zheng and Jiayin Zou and Yilun Chen and Jia Zeng and Tai Wang and Ya-Qin Zhang and Jingjing Liu and Xianyuan Zhan},
booktitle={The Fourteenth International Conference on Learning Representations},
year={2026},
url={https://openreview.net/forum?id=kt51kZH4aG}
}

@inproceedings{xu2024mobility,
  title={Mobility vla: Multimodal instruction navigation with long-context vlms and topological graphs},
  author={Xu, Zhuo and Chiang, Hao-Tien Lewis and Fu, Zipeng and Jacob, Mithun George and Zhang, Tingnan and Lee, Tsang-Wei Edward and Yu, Wenhao and Schenck, Connor and Rendleman, David and Shah, Dhruv and others},
  booktitle={8th Annual Conference on Robot Learning},
  year={2024}
}

@inproceedings{jiang2026wholebodyvla,
title={WholeBody{VLA}: Towards Unified Latent {VLA} for Whole-body Loco-manipulation Control},
author={Haoran Jiang and Jin Chen and Qingwen Bu and Li Chen and Modi Shi and Yanjie Zhang and Delong Li and Chuanzhe Suo and wang chuang and zhihui peng and Hongyang Li},
booktitle={The Fourteenth International Conference on Learning Representations},
year={2026},
url={https://openreview.net/forum?id=OCJmVjyzN7}
}

@inproceedings{rasouli2025distracted,
  title={Distracted Robot: How Visual Clutter Undermine Robotic Manipulation},
  author={Rasouli, Amir and Alban, Montgomery and Pakdamansavoji, Sajjad and Li, Zhiyuan and Zhang, Zhanguang and Wu, Aaron and Zhao, Xuan},
  booktitle={ICRA},
  year={2026}
}

@article{ma2026generalvla,
  title={GeneralVLA: Generalizable Vision-Language-Action Models with Knowledge-Guided Trajectory Planning},
  author={Ma, Guoqing and Wang, Siheng and Zhang, Zeyu and Yu, Shan and Tang, Hao},
  journal={arXiv preprint arXiv:2602.04315},
  year={2026}
}

@article{cutler2015real,
  title={Real-world reinforcement learning via multifidelity simulators},
  author={Cutler, Mark and Walsh, Thomas J and How, Jonathan P},
  journal={IEEE Transactions on Robotics},
  volume={31},
  number={3},
  pages={655--671},
  year={2015},
  publisher={IEEE}
}

@inproceedings{yin2025womap,
title={Wo{MAP}: World Models For Embodied Open-Vocabulary Object Localization},
author={Tenny Yin and Zhiting Mei and Tao Sun and Ola Sho and Anirudha Majumdar and Emily Zhou and Jeremy Bao and Miyu Yamane and Lihan Zha},
booktitle={9th Annual Conference on Robot Learning},
year={2025},
url={https://openreview.net/forum?id=KXzkAje2uQ}
}

@inproceedings{
gao2025adaworld,
title={AdaWorld: Learning Adaptable World Models with Latent Actions},
author={Shenyuan Gao and Siyuan Zhou and Yilun Du and Jun Zhang and Chuang Gan},
booktitle={Forty-second International Conference on Machine Learning},
year={2025},
url={https://openreview.net/forum?id=QQegZj99sk}
}

@inproceedings{
du2025dynaguide,
title={DynaGuide: Steering Diffusion Polices with Active Dynamic Guidance},
author={Max Du and Shuran Song},
booktitle={The Thirty-ninth Annual Conference on Neural Information Processing Systems},
year={2025},
url={https://openreview.net/forum?id=XOw7Yf8qN3}
}

@inproceedings{
huang2025ladiwm,
title={LaDi-{WM}: A Latent Diffusion-Based World Model for Predictive Manipulation},
author={Yuhang Huang and Jiazhao Zhang and Shilong Zou and Xinwang Liu and Ruizhen Hu and Kai Xu},
booktitle={9th Annual Conference on Robot Learning},
year={2025},
url={https://openreview.net/forum?id=o2w2iiMyEU}
}

@inproceedings{
kim2026cosmos,
title={Cosmos Policy: Fine-Tuning Video Models for Visuomotor Control and Planning},
author={Moo Jin Kim and Yihuai Gao and Tsung-Yi Lin and Yen-Chen Lin and Yunhao Ge and Grace Lam and Percy Liang and Shuran Song and Ming-Yu Liu and Chelsea Finn and Jinwei Gu},
booktitle={The Fourteenth International Conference on Learning Representations},
year={2026},
url={https://openreview.net/forum?id=wPEIStHxYH}
}

@inproceedings{
goswami2025osviwm,
title={{OSVI}-{WM}: One-Shot Visual Imitation for Unseen Tasks using World-Model-Guided Trajectory Generation},
author={Raktim Gautam Goswami and Prashanth Krishnamurthy and Yann LeCun and Farshad Khorrami},
booktitle={The Thirty-ninth Annual Conference on Neural Information Processing Systems},
year={2025},
url={https://openreview.net/forum?id=eXO6g7BmOA}
}

@inproceedings{brohan2023rt1,
  title     = {RT-1: Robotics Transformer for Real-World Control at Scale},
  author    = {Brohan, Anthony and Brown, Noah and Carbajal, Justice and Chebotar, Yevgen and 
               Dabis, Joseph and Finn, Chelsea and Gopalakrishnan, Keerthana and Hausman, Karol and 
               Herzog, Alex and Hsu, Jasmine and Ibarz, Julian and Ichter, Brian and Irpan, Alex and 
               Jackson, Tomas and Jesmonth, Sally and Joshi, Nikhil J. and Julian, Ryan and 
               Kalashnikov, Dmitry and Kuang, Yuheng and Leal, Isabel and Lee, Kuang-Huei and 
               Levine, Sergey and Lu, Yao and Malla, Utsav and Manjunath, Deeksha and Mordatch, Igor and 
               Nachum, Ofir and Parada, Carolina and Peralta, Jodilyn and Perez, Emily and Pertsch, Karl and 
               Quiambao, Jornell and Rao, Kanishka and Ryoo, Michael and Salazar, Grecia and 
               Sanketi, Pannag and Sayed, Kevin and Singh, Jaspiar and Sontakke, Sumedh and 
               Stone, Austin and Tan, Clayton and Tran, Huong and Vanhoucke, Vincent and Vega, Steve and 
               Vuong, Quan and Xia, Fei and Xiao, Ted and Xu, Peng and Xu, Sichun and Yu, Tianhe and 
               Zitkovich, Brianna},
  booktitle = {Proceedings of Robotics: Science and Systems (RSS)},
  year      = {2023},
  address   = {Daegu, Republic of Korea},
  month     = {July},
  doi       = {10.15607/RSS.2023.XIX.025}
}

@inproceedings{zitkovich2023rt2,
  title     = {RT-2: Vision-Language-Action Models Transfer Web Knowledge to Robotic Control},
  author    = {Zitkovich, Brianna and Yu, Tianhe and Xu, Sichun and Xu, Peng and Xiao, Ted and
               Xia, Fei and Wu, Jialin and Wohlhart, Paul and Welker, Stefan and Wahid, Ayzaan and
               Vuong, Quan and Vanhoucke, Vincent and Tran, Huong and Soricut, Radu and Singh, Anikait and
               Singh, Jaspiar and Sermanet, Pierre and Sanketi, Pannag and Salazar, Grecia and
               Ryoo, Michael and Reymann, Krista and Rao, Kanishka and Pertsch, Karl and Mordatch, Igor and
               Michalewski, Henryk and Lu, Yao and Levine, Sergey and Lee, Edward Tsang-Wei and
               Lee, Lisa and Leal, Isabel and Kuang, Yuheng and Kalashnikov, Dmitry and Julian, Ryan and
               Joshi, Nikhil and Irpan, Alex and Ichter, Brian and Hsu, Jasmine and Herzog, Alexander and
               Hausman, Karol and Han, Kehang and Gopalakrishnan, Keerthana and Gonzalez Arenas, Montserrat and
               Fu, Chuyuan and Florence, Pete and Finn, Chelsea and Dubey, Avinava and Driess, Danny and
               Ding, Tianli and Choromanski, Krzysztof and Chen, Xi and Chebotar, Yevgen and
               Carbajal, Justice and Brown, Noah and Brohan, Anthony},
  booktitle = {Proceedings of the 7th Conference on Robot Learning (CoRL)},
  series    = {Proceedings of Machine Learning Research},
  volume    = {229},
  pages     = {2165--2183},
  year      = {2023},
  publisher = {PMLR}
}

@inproceedings{belkhale2024rth,
  title     = {RT-H: Action Hierarchies Using Language},
  author    = {Belkhale, Suneel and Ding, Tianli and Xiao, Ted and Sermanet, Pierre and
               Vuong, Quan and Tompson, Jonathan and Chebotar, Yevgen and
               Dwibedi, Debidatta and Sadigh, Dorsa},
  booktitle = {Proceedings of Robotics: Science and Systems (RSS)},
  year      = {2024}
}

@inproceedings{black2025pi0,
  title     = {$\pi_0$: A Vision-Language-Action Flow Model for General Robot Control},
  author    = {Black, Kevin and Brown, Noah and Driess, Danny and Esmail, Adnan and
               Equi, Michael and Finn, Chelsea and Fusai, Niccolo and Groom, Lachy and
               Hausman, Karol and Ichter, Brian and Jakubczak, Szymon and Jones, Tim and
               Ke, Liyiming and Levine, Sergey and Li-Bell, Adrian and Mothukuri, Mohith and
               Nair, Suraj and Pertsch, Karl and Shi, Lucy Xiaoyang and Tanner, James and
               Vuong, Quan and Walling, Anna and Wang, Haohuan and Zhilinsky, Ury},
  booktitle = {Proceedings of Robotics: Science and Systems (RSS)},
  year      = {2025}
}

@inproceedings{ghosh2024octo,
    title={Octo: An Open-Source Generalist Robot Policy},
    author = {{Octo Model Team} and Dibya Ghosh and Homer Walke and Karl Pertsch and Kevin Black and Oier Mees and Sudeep Dasari and Joey Hejna and Charles Xu and Jianlan Luo and Tobias Kreiman and {You Liang} Tan and Lawrence Yunliang Chen and Pannag Sanketi and Quan Vuong and Ted Xiao and Dorsa Sadigh and Chelsea Finn and Sergey Levine},
    booktitle = {Proceedings of Robotics: Science and Systems},
    address  = {Delft, Netherlands},
    year = {2024},
}

@inproceedings{o2024open,
  title={Open x-embodiment: Robotic learning datasets and rt-x models: Open x-embodiment collaboration 0},
  author={O’Neill, Abby and Rehman, Abdul and Maddukuri, Abhiram and Gupta, Abhishek and Padalkar, Abhishek and Lee, Abraham and Pooley, Acorn and Gupta, Agrim and Mandlekar, Ajay and Jain, Ajinkya and others},
  booktitle={2024 IEEE International Conference on Robotics and Automation (ICRA)},
  pages={6892--6903},
  year={2024},
  organization={IEEE}
}

@article{bi2025motus,
  title={Motus: A unified latent action world model},
  author={Bi, Hongzhe and Tan, Hengkai and Xie, Shenghao and Wang, Zeyuan and Huang, Shuhe and Liu, Haitian and Zhao, Ruowen and Feng, Yao and Xiang, Chendong and Rong, Yinze and others},
  journal={arXiv preprint arXiv:2512.13030},
  year={2025}
}

@article{liao2025genie,
  title={Genie envisioner: A unified world foundation platform for robotic manipulation},
  author={Liao, Yue and Zhou, Pengfei and Huang, Siyuan and Yang, Donglin and Chen, Shengcong and Jiang, Yuxin and Hu, Yue and Cai, Jingbin and Liu, Si and Luo, Jianlan and others},
  journal={arXiv preprint arXiv:2508.05635},
  year={2025}
}

@article{li2026causal,
  title={Causal World Modeling for Robot Control},
  author={Li, Lin and Zhang, Qihang and Luo, Yiming and Yang, Shuai and Wang, Ruilin and Han, Fei and Yu, Mingrui and Gao, Zelin and Xue, Nan and Zhu, Xing and others},
  journal={arXiv preprint arXiv:2601.21998},
  year={2026}
}

@misc{ye2026dreamzero,
      title={World Action Models are Zero-shot Policies}, 
      author={Seonghyeon Ye and Yunhao Ge and Kaiyuan Zheng and Shenyuan Gao and Sihyun Yu and George Kurian and Suneel Indupuru and You Liang Tan and Chuning Zhu and Jiannan Xiang and Ayaan Malik and Kyungmin Lee and William Liang and Nadun Ranawaka and Jiasheng Gu and Yinzhen Xu and Guanzhi Wang and Fengyuan Hu and Avnish Narayan and Johan Bjorck and Jing Wang and Gwanghyun Kim and Dantong Niu and Ruijie Zheng and Yuqi Xie and Jimmy Wu and Qi Wang and Ryan Julian and Danfei Xu and Yilun Du and Yevgen Chebotar and Scott Reed and Jan Kautz and Yuke Zhu and Linxi "Jim" Fan and Joel Jang},
      year={2026},
      eprint={2602.15922},
      archivePrefix={arXiv},
      primaryClass={cs.RO},
      url={https://arxiv.org/abs/2602.15922}, 
}

@article{agarwal2025cosmos,
  title={Cosmos world foundation model platform for physical ai},
  author={Agarwal, Niket and Ali, Arslan and Bala, Maciej and Balaji, Yogesh and Barker, Erik and Cai, Tiffany and Chattopadhyay, Prithvijit and Chen, Yongxin and Cui, Yin and Ding, Yifan and others},
  journal={arXiv preprint arXiv:2501.03575},
  year={2025}
}

@article{chen2025robotwin,
  title={Robotwin 2.0: A scalable data generator and benchmark with strong domain randomization for robust bimanual robotic manipulation},
  author={Chen, Tianxing and Chen, Zanxin and Chen, Baijun and Cai, Zijian and Liu, Yibin and Li, Zixuan and Liang, Qiwei and Lin, Xianliang and Ge, Yiheng and Gu, Zhenyu and others},
  journal={arXiv preprint arXiv:2506.18088},
  year={2025}
}

@article{fei25libero-plus,
    title={LIBERO-Plus: In-depth Robustness Analysis of Vision-Language-Action Models},
    author={Senyu Fei and Siyin Wang and Junhao Shi and Zihao Dai and Jikun Cai and Pengfang Qian and Li Ji and Xinzhe He and Shiduo Zhang and Zhaoye Fei and Jinlan Fu and Jingjing Gong and Xipeng Qiu},
    journal = {arXiv preprint arXiv:2510.13626},
    year={2025},
}

@article{yang2026abot,
  title={ABot-M0: VLA Foundation Model for Robotic Manipulation with Action Manifold Learning},
  author={Yang, Yandan and Zeng, Shuang and Lin, Tong and Chang, Xinyuan and Qi, Dekang and Xiao, Junjin and Liu, Haoyun and Chen, Ronghan and Chen, Yuzhi and Huo, Dongjie and others},
  journal={arXiv preprint arXiv:2602.11236},
  year={2026}
}

@inproceedings{
wang2026unified,
title={Unified Vision-Language-Action Model},
author={Yuqi Wang and Xinghang Li and Wenxuan Wang and Junbo Zhang and Yingyan Li and Yuntao Chen and Xinlong Wang and Zhaoxiang Zhang},
booktitle={The Fourteenth International Conference on Learning Representations},
year={2026},
url={https://openreview.net/forum?id=PklMD8PwUy}
}

@INPROCEEDINGS{BuQ-RSS-25, 
    AUTHOR    = {Qingwen Bu AND Yanting Yang AND Jisong Cai AND Shenyuan Gao AND Guanghui Ren AND Maoqing Yao AND Ping Luo AND Hongyang Li}, 
    TITLE     = {{Learning to Act Anywhere with Task-centric Latent Actions}}, 
    BOOKTITLE = {Proceedings of Robotics: Science and Systems}, 
    YEAR      = {2025}, 
    ADDRESS   = {LosAngeles, CA, USA}, 
    MONTH     = {June}, 
    DOI       = {10.15607/RSS.2025.XXI.014} 
}

@article{chen2026h,
  title={H-WM: Robotic Task and Motion Planning Guided by Hierarchical World Model},
  author={Chen, Wenyuan and Huang, Jinbang and Pang, Oscar and Li, Zhiyuan and Hu, Xiao and Zhang, Lingfeng and Zhang, Zhanguang and Coates, Mark and Cao, Tongtong and Quan, Xingyue and Zhang Yingxue},
  journal={arXiv preprint arXiv:2602.11291},
  year={2026}
}

@article{ye2026gigaworld,
  title={GigaWorld-Policy: An Efficient Action-Centered World--Action Model},
  author={Ye, Angen and Wang, Boyuan and Ni, Chaojun and Huang, Guan and Zhao, Guosheng and Li, Hao and Li, Hengtao and Li, Jie and Lv, Jindi and Liu, Jingyu and others},
  journal={arXiv preprint arXiv:2603.17240},
  year={2026}
}

@article{yuan2026fast,
  title={Fast-WAM: Do World Action Models Need Test-time Future Imagination?},
  author={Yuan, Tianyuan and Dong, Zibin and Liu, Yicheng and Zhao, Hang},
  journal={arXiv preprint arXiv:2603.16666},
  year={2026}
}

\newpage

\appendix
\section*{Appendix}
\addcontentsline{toc}{section}{Appendix}

\section{RoboTwin 2.0-Plus Benchmark}
\label{app:robotwin_plus}

To systematically evaluate the robustness of world action models under controlled perturbations, we extend the RoboTwin 2.0~\citep{chen2025robotwin} benchmark with LIBERO-Plus-style~\citep{fei25libero-plus} perturbation categories. We call this extended benchmark \textbf{RoboTwin 2.0-Plus}. It covers \textbf{7 perturbation dimensions} and \textbf{21 sub-dimensions}, applied to all 50 RoboTwin 2.0 dual-arm tasks.

\subsection{Perturbation Taxonomy}

\Cref{tab:perturbation_taxonomy} enumerates all 21 sub-dimensions. Each sub-dimension is individually configurable via YAML flags, enabling both combined and ablation-style evaluation.

\begin{table*}[h]
\caption{\textbf{RoboTwin 2.0-Plus perturbation taxonomy.} All 21 sub-dimensions across 7 LIBERO-Plus dimensions, with implementation-level parameter details.}
\label{tab:perturbation_taxonomy}
\centering
\small
\begin{tabular}{@{}lllp{7.2cm}@{}}
\toprule
\textbf{Dimension} & \textbf{Code} & \textbf{Sub-dimension} & \textbf{Implementation Details} \\ \midrule
\multirow{5}{*}{Sensor Noise}
  & N1 & Motion blur & Gaussian kernel ($r \in [3,15]$, $\sigma \in [1,8]$), rotated by random angle $\in [-30^\circ, 30^\circ]$ \\
  & N2 & Gaussian blur & Isotropic Gaussian ($\sigma \in [1, 10]$) \\
  & N3 & Zoom blur & Multi-scale warp accumulation ($s \in [1.0, 1.56]$, step $\in [0.01, 0.03]$) \\
  & N4 & Fog & Homogeneous white fog with transmittance $e^{-\alpha d}$ ($\alpha \in [0.3, 1.5]$, $d{=}3$) \\
  & N5 & Glass blur & Pixel displacement ($\delta \in [1, 5]$ px) + Gaussian ($\sigma \in [0.5, 2.5]$), 1--3 iterations \\ \midrule
\multirow{4}{*}{Light Conditions}
  & L1 & Diffuse color & Per-channel RGB tint $\in [0.0, 3.5]$ applied to all directional and point lights (always on) \\
  & L2 & Direction & Spherical re-sampling: $\theta \in [8^\circ, 82^\circ]$, $\phi \in [0, 2\pi]$; 35\% chance of dramatic side lighting ($\theta \in [68^\circ, 82^\circ]$). Linked with L4 \\
  & L3 & Specular & Material specular strength $\in [0.3, 6.0]$, shininess $\in [10, 250]$; applied to all scene actors (50\% chance per episode) \\
  & L4 & Shadows & Directional light shadow on/off (50\% chance); co-activated with L2 \\ \midrule
\multirow{3}{*}{Camera Viewpoints}
  & C1 & Distance & Distance scaling $\in [0.85, 1.0]\times$ original (head camera only) \\
  & C2 & Spherical position & Azimuth/elevation perturbation ($\pm 10^\circ$) + $\pm 10\%$ distance variation (disabled by default) \\
  & C3 & Orientation & Yaw/pitch/roll each $\in [0^\circ, 5^\circ]$ with random sign \\ \midrule
Robot Init States & -- & Joint perturbation & Gaussian noise (std${}=0.1$\,rad, clip $\pm 0.225$\,rad) on all joints of both arms; gripper set to extreme (0.05 or 0.95) with $p{=}0.25$ \\ \midrule
\multirow{2}{*}{Background Textures}
  & B1 & Scene theme & Wall + floor RGB color tint, per-channel multiplier $\in [0.4, 1.8]$; combined with upstream texture swap \\
  & B2 & Surface appearance & Table material: metallic $\in [0.0, 0.8]$, roughness $\in [0.05, 0.95]$, per-channel color tint $\in [0.4, 1.8]$ \\ \midrule
\multirow{2}{*}{Objects Layout}
  & O1 & Distractor objects & Variable count per episode (3--15 objects, vs.\ fixed 10 in upstream) \\
  & O2 & Target pose & Position: Gaussian noise ($\sigma{=}2$\,cm, x/y only); orientation: uniform yaw $\pm 15^\circ$ \\ \midrule
\multirow{3}{*}{Language Instructions}
  & R1 & Distraction & Irrelevant conversational wrapping (${\sim}30\%$ of combined episodes) \\
  & R2 & Common-sense rewording & Object names $\to$ functional descriptions, verb synonyms (${\sim}50\%$) \\
  & R3 & Reasoning chain & Imperative $\to$ goal-state / outcome description (${\sim}20\%$) \\
\midrule
\multicolumn{4}{@{}l}{\textit{Total: 7 dimensions, 21 sub-dimensions implemented; 20/21 active by default (C2 disabled to avoid instability)}} \\
\bottomrule
\end{tabular}
\end{table*}

\subsection{Evaluation Protocol}

A full RoboTwin 2.0-Plus evaluation requires \textbf{8 configs} per task: 1 clean baseline + 7 perturbation branches (\Cref{tab:eval_configs}). Each branch activates exactly one perturbation dimension while keeping all others at their default (clean) values, isolating the effect of each dimension on policy performance. Each config runs 50 episodes per task.

\begin{table}[h]
\caption{\textbf{RoboTwin-Plus evaluation configs.} The 8 required configs for a complete robustness evaluation.}
\label{tab:eval_configs}
\centering
\small
\begin{tabular}{@{}clll@{}}
\toprule
\textbf{\#} & \textbf{Config} & \textbf{Perturbation} & \textbf{Active Sub-dims} \\ \midrule
0 & \texttt{demo\_clean} & None (clean baseline) & -- \\
1 & \texttt{demo\_vision\_noise} & Sensor Noise & N1--N5 (cycled) \\
2 & \texttt{demo\_light} & Lighting & L1 (always) + L2/L3/L4 (stochastic) \\
3 & \texttt{demo\_camera} & Camera Viewpoints & C1 + C3 (C2 disabled by default) \\
4 & \texttt{demo\_robot\_state} & Robot Initial State & Joint noise + gripper extremes \\
5 & \texttt{demo\_background\_plus} & Background & B1 (texture + color tint) + B2 \\
6 & \texttt{demo\_objects\_plus} & Objects Layout & O1 (variable count) + O2 (pose) \\
7 & \texttt{demo\_language\_plus} & Language & R1 + R2 + R3 (combined) \\
\bottomrule
\end{tabular}
\end{table}

\subsection{Sub-dimension Details}

\paragraph{Sensor Noise (N1--N5).} One of five noise types is applied per episode, cycled deterministically by episode index: \textbf{N1} motion blur (oriented Gaussian kernel), \textbf{N2} Gaussian blur (isotropic), \textbf{N3} zoom blur (multi-scale warp accumulation around the image center), \textbf{N4} fog (homogeneous white fog with exponential transmittance), and \textbf{N5} glass blur (random pixel displacement with iterative Gaussian smoothing). Noise severity is sampled uniformly in $[2, 3]$ per episode. All corruptions are applied to RGB observations at render time for both head and wrist cameras.

\paragraph{Lighting.} \textbf{L1} (diffuse color): a strong per-channel RGB tint $\in [0.0, 3.5]$ is sampled and applied to all directional and point light sources; always active. \textbf{L2} (direction): the primary directional light is re-oriented via spherical coordinates ($\theta$, $\phi$), with a 35\% chance of dramatic side-lighting ($\theta \in [68^\circ, 82^\circ]$); activated jointly with \textbf{L4} (shadows): directional light shadow casting is toggled on/off (50\% per episode). \textbf{L3} (specular): specular strength $\in [0.3, 6.0]$ and shininess $\in [10, 250]$ are applied to all scene actor materials (50\% chance per episode).

\paragraph{Camera Viewpoints.} Applied to the head camera only. \textbf{C1} (distance): viewing distance is scaled by a factor $\in [0.85, 1.0]$. \textbf{C2} (spherical position): azimuth and elevation are perturbed by up to $\pm 10^\circ$ with $\pm 10\%$ distance variation; disabled by default in the evaluation config but can be enabled via a YAML flag for ablation. \textbf{C3} (orientation): yaw, pitch, and roll are independently perturbed in $[0^\circ, 5^\circ]$ with random sign.

\paragraph{Robot Initial States.} Gaussian noise with $\text{std}=0.1$\,rad (clipped to $\pm 0.225$\,rad per joint) is added to the home joint angles of both arms. Additionally, each gripper is set to an extreme position (0.05 or 0.95) with probability 0.25.

\paragraph{Background.} \textbf{B1} (scene theme): in addition to the upstream texture swap, a random per-channel RGB multiplier $\in [0.4, 1.8]$ is applied to the wall color, and the floor receives an independent random color. \textbf{B2} (surface appearance): the table material is randomized with metallic $\in [0.0, 0.8]$, roughness $\in [0.05, 0.95]$, and a per-channel color tint $\in [0.4, 1.8]$.

\paragraph{Objects Layout.} \textbf{O1} (distractor objects): the number of task-irrelevant objects placed on the table is sampled uniformly from $[3, 15]$ per episode, replacing the upstream fixed count of 10. \textbf{O2} (target pose): after scene initialization, task-relevant objects receive Gaussian position noise ($\sigma = 2$\,cm in $x$ and $y$; $z$ unchanged) and uniform yaw rotation up to $\pm 15^\circ$.

\paragraph{Language Instructions.} 2{,}500 pre-generated instruction variants (50 tasks $\times$ 50 variants) are stored as JSON files. In the combined config, each episode samples one variant: \textbf{R1} (distraction, ${\sim}30\%$) wraps the original instruction in irrelevant conversational context; \textbf{R2} (common-sense rewording, ${\sim}50\%$) replaces object names with functional descriptions and uses verb synonyms; \textbf{R3} (reasoning chain, ${\sim}20\%$) rewrites the imperative instruction as a goal-state or outcome description. Individual R2-only and R3-only configs are provided for ablation studies.

\section{Benchmark comparison}

\label{app:bench_comp}
\begin{table}[h]
\caption{\textbf{Benchmark comparison} between LIBERO-Plus and RoboTwin 2.0-Plus.}
\label{tab:benchmark_comparison}
\centering
\begin{tabular}{@{}lll@{}}
\toprule
Aspect & LIBERO-Plus & RoboTwin 2.0-Plus \\ \midrule
Simulator & MuJoCo (robosuite) & SAPIEN (ManiSkill3) \\
Robot & Franka Panda (7-DoF) & Aloha-AgileX (14-DoF) \\
Arms & Single arm & Dual arm (bimanual) \\
Cameras & 2 (third-person + wrist) & 3 (head + 2 wrist) \\
Image resolution & $256\times256$ & $320\times240$ \\
Native action space & 7-dim delta EEF & 14-dim joint positions \\
Control mode & OSC (delta EE pose) & Joint position control \\
Control frequency & 10\,Hz & 25--30\,Hz \\
Base tasks & 40 (4 suites $\times$ 10) & 50 collaborative tasks \\
Training demos & 50 per task & 50 clean + 500 randomized \\
Total trajectories & 22,400 & 27,500 \\
Distractor objects & 416 & 731 (147 categories) \\
\bottomrule
\end{tabular}
\end{table}

\newpage

\section{Comparison of Evaluated Models} \label{app:model_comp}

\begin{table}[h]
\caption{\textbf{Model setup comparison.} Action representation, dimensionality, chunk size, and control frequency for each evaluated model. Entries marked with ``--'' indicate values inherited from the base model or not explicitly specified. Unless noted, models on LIBERO follow the native axis-angle rotation convention.}
\label{tab:model_setup}
\centering
\begin{tabular}{@{}lllll@{}}
\toprule
Model & Action Repr. & Dim & Chunk & Freq \\ \midrule
\multicolumn{5}{@{}l}{\textit{VLA Models}} \\
Pi0 & Delta EEF & 7 & 50 & 50\,Hz \\
Pi0-FAST & Delta EEF (FAST tokenized) & 7 & 50 & 15--50\,Hz \\
Pi0.5 & Absolute EEF / joints & 18--19 & 50 & 50\,Hz \\
OpenVLA-OFT & Delta EEF & 7 & 8 & 3--10\,Hz \\
UniVLA & Delta EEF (DCT tokenized) & 7 & 10 & -- \\
RIPT-VLA & Delta EEF (inherited) & 7 & 8 & -- \\
xVLA & Absolute EEF (6D rot.) & 10/20 & 32 & -- \\
VLA-JEPA & Delta EEF & 7 & 7 & -- \\
MOTUS & Latent (optical flow) & 14 & 48 & 30\,Hz \\ \midrule
\multicolumn{5}{@{}l}{\textit{World Action Models}} \\
GE-Act & Absolute EEF (Euler) & 7/14 & 54 & 30\,Hz \\
Cosmos-Policy & Latent-frame + native & 7/14 & 16 & -- \\
LingBot-VA & Absolute EEF (quat.) + joints & 30 & 4 & 50\,Hz \\
\bottomrule
\end{tabular}
\end{table}

We evaluate both VLA models and WAMs. The action representation and control details of each model are summarized in \Cref{tab:model_setup}. On the VLA side, \textbf{Pi0}~\cite{black2025pi0} and \textbf{Pi0-FAST} use delta end-effector control with axis-angle rotation and generate 50-step action chunks via flow matching and FAST tokenization, respectively. \textbf{Pi0.5}~\citep{pi05} extends Pi0 to predict absolute target poses with 50-step chunks at 50\,Hz, and supports both end-effector and joint-space control depending on the platform. \textbf{OpenVLA-OFT}~\citep{openvla2025} follows the native 7-dim delta end-effector convention on LIBERO with an action chunk size of 8, and \textbf{RIPT-VLA} inherits its action space from the OpenVLA-OFT base model. \textbf{UniVLA} tokenizes native actions via DCT (FAST tokenizer) with a chunk size of 10 on LIBERO. \textbf{xVLA}~\citep{zheng2026xvla} is notable for employing a unified absolute end-effector representation with 6D rotation across all benchmarks, outputting 10-dim per arm (padded to 20-dim for dual-arm), with a chunk size of 32. \textbf{VLA-JEPA}~\citep{sun2026vla} predicts delta end-effector actions with axis-angle rotation using a conditional flow-matching action head and a prediction horizon of 7 steps.

On the WAM side, \textbf{Cosmos-Policy}~\citep{kim2026cosmos} encodes actions as latent frames within its video diffusion process, using the native 7-dim action space on LIBERO with a chunk size of 16. \textbf{GE-Act}~\citep{liao2025genie} predicts absolute end-effector poses with Euler angle rotation and generates 54-step action trajectories via a lightweight flow-matching decoder at 30\,Hz. \textbf{MOTUS}~\citep{bi2025motus} takes a fundamentally different approach by operating in a learned 14-dimensional latent action space derived from optical flow, which is mapped to robot-specific commands during fine-tuning, with a chunk size of 48 at 30\,Hz. \textbf{LingBot-VA}~\citep{li2026causal} uses a 30-dimensional representation combining absolute end-effector poses (with quaternion rotation) and joint angles for both arms, with a compact chunk size of 4 at 50\,Hz.

\subsection{Detailed LIBERO-Plus Results} \label{app:detailed_libero_plus}

\begin{table}[t]
\centering
\caption{Detailed LIBERO-Plus results for each task suite. \textbf{Lang.}: Language. \textbf{BG}: Background. The Fast-WAM LIBERO checkpoint was trained on clean demonstrations only, whereas its RoboTwin checkpoint was trained on clean + domain-randomized demonstrations (cf.\ \textbf{RQ 2}).}
\label{tab:libero_plus_detail}
\begin{tabular}{@{}L{1.5cm} L{2.5cm} C{1.0cm} C{1.0cm} C{1.0cm} C{1.0cm} C{1.0cm} C{1.0cm} C{1.0cm} C{1.0cm}@{}}
\toprule
Category & Model & Camera & Robot & Lang. & Light & BG & Noise & Layout & Total \\ \midrule
\multirow{7}{*}{Spatial} & $\pi_0$ & 70.7 & 49.1 & 67.9 & 92.8 & 95.0 & 87.7 & 94.0 & 78.6 \\
 & $\pi_{0.5}$ & 76.9 & 85.4 & 90.3 & 97.6 & 95.3 & 91.7 & 97.7 & 90.3 \\
 & X-VLA & 24.5 & 91.4 & 82.1 & 100.0 & 100.0 & 63.5 & 93.8 & 77.7 \\
 & VLA-JEPA & 74.2 & 71.7 & 96.1 & 97.9 & 99.6 & 77.8 & 95.3 & 86.9 \\
 & GE-Act & 72.3 & 75.7 & 90.8 & 100.0 & 89.1 & 97.7 & 89.6 & 87.5 \\
 & Cosmos-policy & 81.1 & 59.7 & 88.2 & 89.0 & 91.1 & 98.0 & 99.2 & 86.6 \\
 & Fast-WAM & 14.4 & 44.0 & 69.5 & 87.3 & 69.8 & 35.0 & 60.5 & 54.4 \\ \midrule
\multirow{7}{*}{Object} & $\pi_0$  & 80.1 & 31.9 & 75.4 & 94.3 & 85.9 & 87.9 & 76.2 & 74.7\\
 & $\pi_{0.5}$ & 91.4 & 70.6 & 92.4 & 98.7 & 98.0 & 95.5 & 88.6 & 90.0 \\
 & X-VLA & 19.7 & 97.2 & 90.7 & 92.6 & 98.4 & 51.4 & 55.6 & 69.3 \\
 & VLA-JEPA & 69.2 & 57.8 & 90.8 & 100.0 & 100.0 & 75.1 & 87.3 & 81.0 \\
 & GE-Act & 81.3 & 83.4 & 83.6 & 98.3 & 94.8 & 97.2 & 79.7 & 87.7 \\
 & Cosmos-policy & 88.4 & 61.6 & 81.9 & 100.0 & 96.4 & 94.3 & 87.1 & 86.2 \\
 & Fast-WAM & 25.3 & 64.1 & 96.3 & 97.9 & 77.8 & 63.7 & 73.0 & 71.2 \\ \midrule
\multirow{7}{*}{Goal} & $\pi_0$ & 56.6 & 43.3 & 43.2 & 90.3 & 84.7 & 82.8 & 59.8 & 63.4 \\
 & $\pi_{0.5}$ & 80.1 & 80.2 & 71.5 & 98.6 & 92.9 & 94.2 & 71.1 & 82.7 \\
 & X-VLA & 27.2 & 90.2 & 58.8 & 95.7 & 97.9 & 67.5 & 61.2 & 68.7 \\
 & VLA-JEPA & 72.3 & 73.1 & 75.3 & 89.9 & 94.3 & 69.9 & 68.7 & 76.3 \\
 & GE-Act & 52.5 & 71.9 & 46.3 & 93.9 & 87.2 & 83.4 & 68.9 & 70.0 \\
 & Cosmos-policy & 75.7 & 64.1 & 67.6 & 99.6 & 95.0 & 91.6 & 56.2 & 76.4 \\
 & Fast-WAM & 8.1 & 24.7 & 49.8 & 75.3 & 44.5 & 23.7 & 48.0 & 39.2 \\ \midrule
\multirow{7}{*}{Long} & $\pi_0$  & 38.7 & 39.9 & 69.7 & 79.2 & 72.3 & 64.6 & 77.6 & 61.3 \\
 & $\pi_{0.5}$ & 54.2 & 74.8 & 89.8 & 92.7 & 92.7 & 78.8 & 87.2 & 79.9 \\
 & X-VLA & 22.2 & 79.9 & 71.3 & 95.6 & 94.5 & 68.2 & 76.6 & 69.9 \\
 & VLA-JEPA & 42.5 & 68.5 & 90.9 & 78.4 & 81.3 & 44.3 & 86.2 & 67.9 \\
 & GE-Act & 38.9 & 77.1 & 91.4 & 90.5 & 74.4 & 86.0 & 84.6 & 76.6  \\
 & Cosmos-policy & 59.2 & 67.4 & 89.8 & 97.4 & 74.7 & 88.0 & 91.7 & 80.2 \\
 & Fast-WAM & 17.7 & 45.3 & 60.1 & 52.2 & 22.8 & 28.5 & 61.2 & 41.1 \\ \bottomrule
\end{tabular}
\end{table}

\end{document}